\pdfoutput=1

\documentclass[11pt]{article}

\usepackage[final]{acl}
\usepackage{times}
\usepackage{latexsym}

\usepackage[T1]{fontenc}

\usepackage[utf8]{inputenc}

\usepackage{microtype}

\usepackage{inconsolata}

\usepackage{graphicx}

\usepackage{tikz}
\usetikzlibrary{shapes.multipart,calc,arrows,fit,positioning}
\usepackage[inline]{enumitem}
\usepackage{subcaption}
\usepackage{tikzscale}  
\usepackage{multirow}
\usepackage{makecell}
\usepackage{arydshln}
\usepackage{tabularx}

\makeatletter
\newcommand{\inst}[1]{\unskip$^{#1}$}
\makeatother

\title{Can Rule-Based Insights Enhance LLMs for Radiology Report Classification? Introducing the RadPrompt Methodology.}

\author{ 
\textbf{Panagiotis Fytas \inst{1} \quad}
\textbf{Anna Breger \inst{*,2,3} \quad}
\textbf{Ian Selby \inst{*,4,5}}
\\
\textbf{Simon Baker \inst{1} \quad}
\textbf{Shahab Shahipasand \inst{5} \quad}
\textbf{Anna Korhonen \inst{1} }
\\
\\
\inst{1}Language Technology Lab, University of Cambridge \\
\inst{2}Department of Applied Mathematics and Theoretical Physics, University of Cambridge\\
\inst{3}Center of Medical Physics and Biomedical Engineering, Medical University of Vienna\\
\inst{4}Department of Radiology, University of Cambridge\\
 \inst{5}Cambridge University Hospitals, NHS Foundation Trust \\
\texttt{pf376@cam.ac.uk}
}

\begin{document}
\maketitle
\begin{abstract}
Developing imaging models capable of detecting pathologies from chest X-rays can be cost and time-prohibitive for large datasets as it requires supervision to attain state-of-the-art performance. Instead, labels extracted from radiology reports may serve as distant supervision since these are routinely generated as part of clinical practice. Despite their widespread use, current rule-based methods for label extraction rely on extensive rule sets that are limited in their robustness to syntactic variability. To alleviate these limitations, we introduce RadPert, a rule-based system that integrates an uncertainty-aware information schema with a streamlined set of rules, enhancing performance. Additionally, we have developed RadPrompt, a multi-turn prompting strategy that leverages RadPert to bolster the zero-shot predictive capabilities of large language models, achieving a statistically significant improvement in weighted average F1 score over GPT-4 Turbo. Most notably, RadPrompt surpasses both its underlying models, showcasing the synergistic potential of LLMs with rule-based models. We have evaluated our methods on two English Corpora: the MIMIC-CXR gold-standard test set and a gold-standard dataset collected from the Cambridge University Hospitals.

\end{abstract}

\section{Introduction}

\def\thefootnote{*}\footnotetext{Equal contribution.}\def\thefootnote{\arabic{footnote}}

Supervised deep learning for medical imaging classification has accomplished significant milestones. In the chest X-ray (CXR) domain, such models have exhibited predictive capabilities on par with expert physicians \citep{rajpurkar2018deep, tang2020automated} and are being utilized in collaborative settings to increase clinician accuracy \citep{rajpurkar2020chexaid}.

Annotating medical images, however, is expensive and arduous: it requires a committee of expert radiologists to resolve the inherently high degree of annotator variance and subjectivity \citep{razzak2018deep}. This issue is particularly problematic considering the global shortage of radiologists \citep{jeganathan2023growing, kalidindi2023workforce, konstantinidis2023shortage}. Instead, we often have access to a form of distant supervision: the radiology report. Radiology reports are semi-structured free-text interpretations of an X-ray image and are generated as a routine part of clinical practice to communicate findings.

In the past, rule-based models \citep{chexpert, negbio} have been used to extract structured labels from radiology reports in various imaging datasets, including ChestX-ray14 \citep{chestxray14}, CheXpert \citep{chexpert}, MIMIC-CXR \citep{johnson2019mimic} and BRAX \citep{reis2022brax}. However, those rule-based methods are often based on elementary techniques and, thus, exhibit limited robustness to syntactic variation. Naturally, supervised deep learning models offer superior performance through their robustness to syntactic variability \citep{smit2020chexbert, visualchexbert}. In contrast, Large Language Models (LLMs) represent a significant improvement over rule-based models in an unsupervised setting and have achieved impressive performance in the field of radiology \citep{infante2024large, adams2023leveraging, liu2023exploring}.

In this paper, we present RadPert, a rule-based model built on the RadGraph knowledge graph \citep{RadGraphNeurips}. RadPert leverages entity-level uncertainty labels from RadGraph, reducing the need for a comprehensive rule set and enhancing its resilience to syntactic variations. We have evaluated RadPert internally on MIMIC-CXR and externally on a dataset collected from the Cambridge University Hospitals (CUH). RadPert surpasses CheXpert, the former rule-based state-of-the-art (SOTA), by achieving statistically significant improvement in weighted average F1 score. 

\begin{figure*}
    \centering
    \includegraphics[scale=0.7]{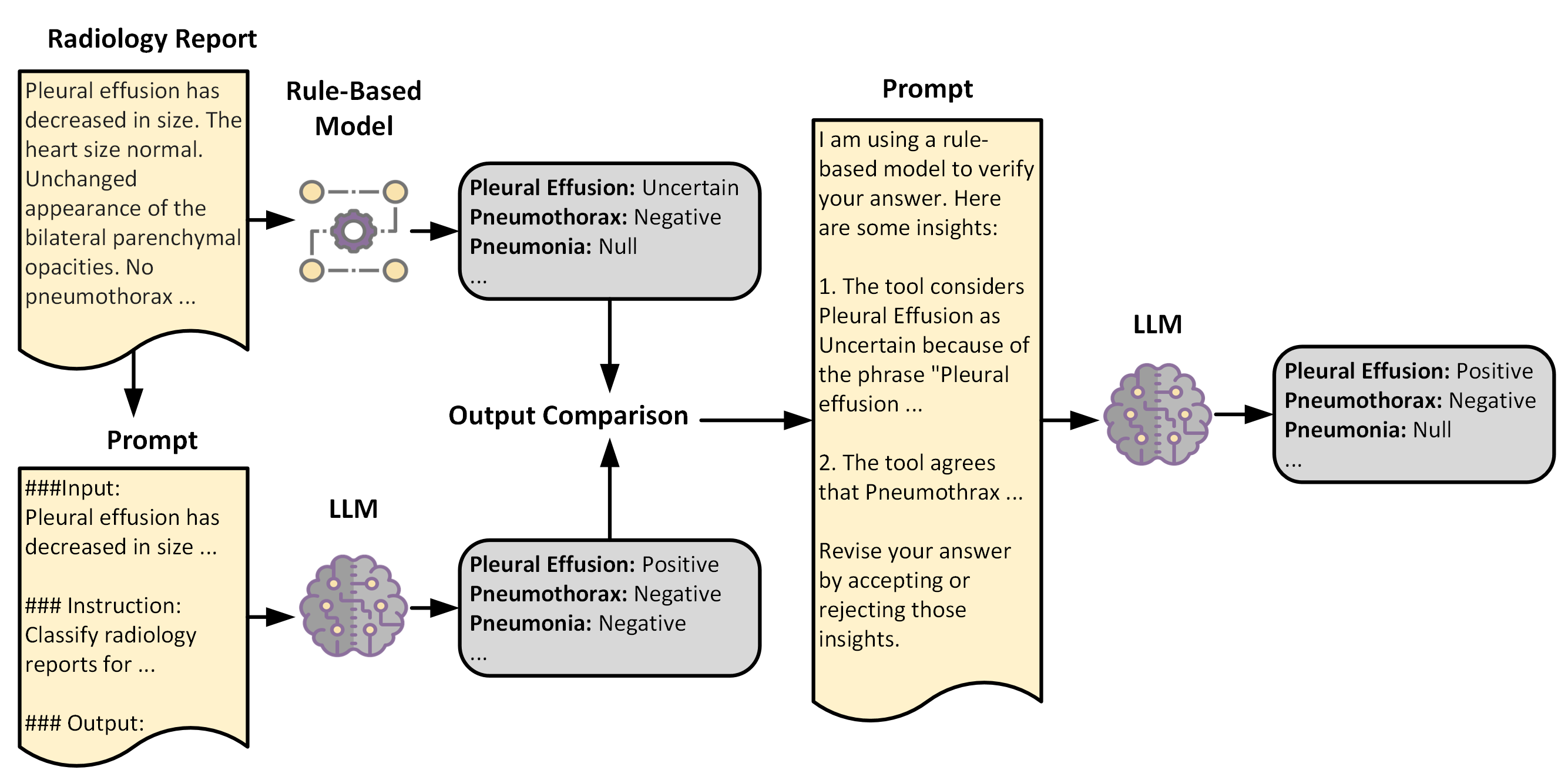}
    \caption{Overview of the RadPrompt methodology. RadPrompt utilizes the rule-based RadPert model to detect potential errors in the original (first-turn) LLM classification decision. A second-turn prompt is then constructed, offering evidence that may cause the LLM to revise its original classification outcome.}
    \label{fig:radprompt}
\end{figure*}

Furthermore, we explore the collaborative potential of LLMs with rule-based models through RadPrompt. RadPrompt is a multi-turn prompting strategy that employs RadPert as an implicit means of encoding medical knowledge (Figure \ref{fig:radprompt}). In fact, RadPrompt, based on GPT-4 Turbo, manages to outperform both its underlying models in a zero-shot setting.



\section{Related Work}


Numerous natural language processing methods have been developed to derive structured predictions from radiology reports \citep{negbio, hassanpour2017performance, pons2016natural, bozkurt2019automated, wang2018clinical}. Many of those approaches are designed for the multitask classification of radiology reports, written in English, into labels representing prevalent pathologies from CXRs. Each such label can exhibit one of four output classes: \textit{Null, Positive, Negative} and \textit{Uncertain}. CheXpert \citep{chexpert}, the rule-based SOTA, follows an approach based on regular expression matching and the Universal Dependency Graph (UDG) of a radiology report.  Due to the rudimentary regular expression matching, however, CheXpert is sensitive to syntactic variation. Thus, multiple over-generalized rules are used in an attempt to alleviate these shortcomings. Furthermore, the UDG is a type of information extraction that does not explicitly identify negation and uncertainty. Therefore, its ability to detect uncertainty in complex phrases is hampered despite the extensive rule set. Extensions of CheXpert have been developed for Brazilian Portuguese \citep{reis2022brax} and German \citep{wollek2024german}. CheXbert \citep{smit2020chexbert} is a semi-supervised model pre-trained on automatically extracted labels from the CheXpert model, fine-tuned on manually annotated reports, and evaluated on 687 MIMIC-CXR gold-standard test set reports. However, the published model weights\footnote{\href{https://github.com/stanfordmlgroup/CheXbert}{https://github.com/stanfordmlgroup/CheXbert}} of CheXbert differ from the original model. This discrepancy complicates comparisons on the MIMIC-CXR dataset as the published model is fine-tuned on unspecified MIMIC-CXR manually annotated reports, which can potentially overlap with the MIMIC-CXR gold-standard test set. 

Recent work has also explored the adoption of LLMs for radiology report classification. Specifically, \citet{dorfner2024open} examine the zero and few-shot capabilities of LLMs. However, they mainly treat the task as a binary classification for each pathology. Namely, for multitask classification, they only report the few-shot results on an unpublished institutional dataset. CheX-GPT \citep{gu2024chex} utilizes zero-shot GPT-4 labels as a distant supervision to fine-tune a BERT-based model. Nonetheless, they also simplify the task into binary classification. 



Alternative approaches to the classification of chest X-rays (CXRs) explore moving away from the distantly supervised paradigm of training unimodal vision models on classifying structured labels extracted from radiology reports. In lieu of structured prediction, Vision-Language (VL) models are trained to align the embedding representations of CXRs with the representations of the corresponding radiology reports via self-supervised contrastive learning objectives \citep{huang2021gloria, boecking2022making, tiu2022expert, wang2022medclip, bannur2023learning}. This alignment task is transformed into CXR classification through the cosine similarity of CXR embeddings to the embeddings of textual prompts representing the existence or absence of pathologies. However, vision models trained with the structured prediction paradigm outperform VL models such as CheXzero \citep{tiu2022expert}, even when the latter utilizes an expert-annotated validation set for selecting optimal classification thresholds.

In this paper, we will focus on improving the unsupervised SOTA for the multitask classification of radiology reports.

\section{Methods}

\subsection{Task}

Similar to CheXpert and CheXbert, we will focus on the multitask classification of CXR radiology reports. Specifically, our models classify thirteen labels that correspond to pathologies {(Atelectasis, Edema, Cardiomegaly, Consolidation, Enlarged Cardiomediastinum, Fracture, Lung Lesion, Lung Opacity, Pleural Effusion, Pleural Other, Pneumothorax, Support Devices and Pneumonia)}, with each label having four possible output classes: \textit{Null, Positive, Negative} and {Uncertain}.
A pathology is classified as \textit{Null} if there are no references to it in the radiology report. It is considered \textit{Negative} when its absence is explicitly mentioned. \textit{Positive} classes entail that the existence of the corresponding pathology is specified in the report. Finally, \textit{Uncertain} classes imply that while the pathology is discussed in the report, its existence cannot be determined.


\subsection{RadPert}

In order to overcome the limitations of existing tools, we have designed RadPert. RadPert incorporates hand-crafted rules with the RadGraph \citep{RadGraphNeurips} knowledge graph.


\subsubsection{RadGraph Information Schema}

RadGraph \citep{RadGraphNeurips} defines an information schema specifically designed for radiology reports. It contains two top-level entity types: \textit{{Anatomy} (ANAT) }and \textit{{Observation} (OBS)}. \textit{{Anatomy}} entities describe bodily anatomical structures (e.g. ``lobe'') and their spatial characteristics (e.g. ``left''). \textit{{Observation}} entities include pathological abnormalities (e.g. ``opacities''), diagnosed diseases (e.g. ``pneumonia'') and various other characteristics (e.g. ``acute''). It is important to note that \textit{{Observation}} entities are further categorized into three second-level attributes: \textit{{Definitely Present} (DP)}, \textit{{Definitely Absent} (DA)} and \textit{{Uncertain} (U)}.

Additionally, RadGraph defines three types of directed relations between entities. Firstly, the \textit{suggestive of} relation indicates that some \textit{{Observation}} implies the existence of another \textit{{Observation}}. Secondly, \textit{located at} relations account for \textit{{Observations}} relating to specific \textit{{Anatomies}}. Finally, \textit{modify} relations can exist only between the same type of entity and describe the characteristics relating to a specific entity (e.g., \textit{modify}(``left'', ``lung'')). 

The RadGraph model is based on the DyGIE++ \citep{wadden2019entity} framework initialized with PubMedBERT weights \citep{gu2021domain}. The model is fine-tuned on 500 expert-annotated MIMIC-CXR reports based on the RadGraph information schema.

\subsubsection{RadPert Pipeline}

\begin{figure*}
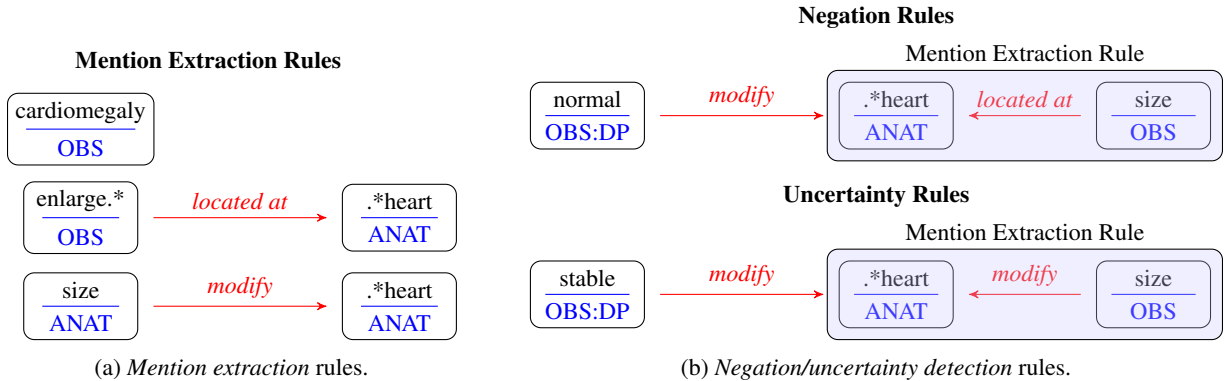

    \centering
    \begin{subfigure}{.37\textwidth}
        \centering
        \includegraphics[width=\linewidth]{fig/cardiomegaly_mention.tikz}
        \caption{\textit{Mention extraction} rules.}
        \label{subfig:radpert-mention}
    \end{subfigure}
    \hfill
    \begin{subfigure}{.57\textwidth}
        \centering
        \includegraphics[width=\linewidth]{fig/cardiomegaly_uncneg.tikz}
        \caption{\textit{Negation/uncertainty detection} rules.}
        \label{subfig:radpert-unc-neg}
    \end{subfigure}
    \caption{Examples of RadPert rules for Cardiomegaly. The rules take the form of graphs that follow the RadGraph \citep{RadGraphNeurips} information schema. The ``.*'' symbolizes allowing the matching of different prefixes and suffixes within the entity span.}
\end{figure*}

RadPert employs the following four-stage pipeline:

\textbf{Knowledge graph extraction.} We first extract the RadGraph entities and relations from radiology reports. Utilizing RadGraph instead of the UDG allows uncertainty and negation classes to be extracted at an entity level. Thus, the negation and the uncertainty of various complex phrases can be determined based on those classes, reducing the need for complex negation and uncertainty rules.

\textbf{Mention extraction}. In this stage, for each pathology label, we have adapted and simplified the CheXpert rules \citep{chexpert} so they can be applied to RadGraph entities and relations. Essentially, those rules can be represented as graphs based on the RadGraph information schema. Figure \ref{subfig:radpert-mention} includes examples of mention extraction rules in the form of graphs. Checking whether a pathology is mentioned in a radiology report amounts to determining whether any rule-graphs for the specific pathology are subgraphs of the radiology report knowledge graph\footnote{ This problem corresponds to the edge-colored and node-colored variant of Induced Subgraph Isomorphism. Exhaustive search with subgraphs of fixed-length has polynomial complexity \citep{FLODERUS2015119}.}. If none of the pathology rules match a given radiology report, then the class for that pathology is \textit{Null}.

\textbf{Negation/uncertainty detection.} We next aim to determine whether an extracted mention is \textit{Positive}, \textit{Negative}, or \textit{Uncertain}. 
For mentions that contain \textit{Observation} entities in their subgraph, the uncertainty quantifier of the \textit{Observation} determines the initial class of that mention.
For instance, if a ``heart'' \textit{Anatomy} is connected with an ``enlarged'' \textit{Observation}, which is characterized as \textit{Definitely Absent}, then that mention will be labeled as \textit{Negative}. If a mention only possesses \textit{Anatomy} entities, then we consider by default that mention to be \textit{Positive}. However, certain phrases contain implicit negations/uncertainties. In cases such as ``normal heart size'', the entity ``normal'' would be considered under RadGraph a \textit{Definitely Present} \textit{Observation} attached to an \textit{Anatomy}. Thus, in order to detect such implicit negations/uncertainties and determine the final uncertainty class for each pathology, we have developed a negation and an uncertainty rule set. Both rule sets are constructed from hand-crafted rules in the form of graphs. Examples of Cardiomegaly negation/uncertainty rules can be observed in Figure \ref{subfig:radpert-unc-neg}. When a negation rule is activated, the initial class of the mention will be negated (i.e., \textit{Positive} becomes \textit{Negative} and \textit{Negative} becomes \textit{Positive}). However, when an uncertainty rule is matched, RadPert considers the class of the mention to be \textit{Uncertain}.

\textbf{Mention aggregation.} After extracting and classifying all mentions in a radiology report for a specific label, RadPert aggregates them into the final uncertainty class for that label. Similarly to CheXpert \citep{chexpert}, we prioritize positive mentions, followed by uncertain ones, while negative mentions have the lowest priority.


\subsection{RadPrompt}

RadPert, through its rules, implicitly encodes expert knowledge vital to classifying radiology reports. However, as a rule-based system,  it is still sensitive to syntactic and lexical variability. To alleviate this limitation, we propose RadPrompt, a zero-shot prompting technique that injects prompts with insights derived from the application of RadPert.
RadPrompt, as seen in Figure \ref{fig:radprompt}, employs a two-turn prompting strategy.

In the first turn, the zero-shot prompt contains instructions, which define the task, and the radiology report that needs to be classified.
After a response is received from the LLM, the first-turn classification outcome is compared with the output of RadPert.

In the second turn, a prompt is constructed by specifying that a rule-based model is used to verify the validity of the LLM's answer. Hints are then added by specifying for each pathology either RadPert's agreement with the LLM or the radiology report sentence that leads RadPert to a disagreement. This is possible since RadPert, as a rule-based system, allows the detection of the specific mention that leads to the classification decision. Finally, the prompt instructs the LLM to adjust its answer by accepting or rejecting RadPert's hints. In Table \ref{tab:prompt} of the Appendix, we present the format of our first and second-turn prompts.

\subsubsection{Base Model}

As a base model for the RadPrompt strategy, we explore various LLMs, including API-based models such as Gemini-1.5 Pro \citep{reid2024gemini}, Claude-3 Sonnet, GPT-4 Turbo \citep{achiam2023gpt}, and Llama-2 \citep{touvron2023llama}. In the case of Llama-2, we are using the 70 billion parameter chat variant, quantized with the Int 4 AWQ method \citep{lin2023awq}, which we run locally with a single NVIDIA RTX 6000 Ada GPU. 


\section{Results and Discussion}

\subsection{Evaluation}

To allow comparison with previous work \citep{chexpert, smit2020chexbert}, for each pathology, we evaluate our methodology based on the weighted average F1 score across three aspects of the task: negation detection, positive mention detection, and uncertainty detection. We report the F1 scores of the sub-tasks in the Appendix. Each of those sub-tasks amounts to binary classification. For instance, \textit{Negative} classes are transformed into positive in negation detection, while the other classes are transformed into negative. Positive mention detection and uncertainty detection are constructed with an analogous logic. The reported scores correspond to the averages across 1000 bootstrap replicates \citep{efron1986bootstrap}, reported along the $95\%$ Confidence Intervals (CI). 

\subsection{Data}

For internal evaluation, we are evaluating the models on the gold-standard test set of annotated radiology reports used in the MIMIC-CXR paper \citep{johnson2019mimic}. MIMIC-CXR is considered an internal dataset for methods based on RadPert since RadGraph is trained on MIMIC-CXR radiology reports. The MIMIC-CXR gold-standard test set contains 687 radiology reports that do not overlap with the training and validation set of RadGraph.

For external evaluation, we have collected a private dataset from the Cambridge University Hospitals in Cambridge, UK. The CUH dataset consists of 650 radiology reports annotated by a single consultant radiologist with six years of experience, using the same annotation guidelines as MIMIC-CXR\footnote{MIMIC-CXR annotation guidelines were provided upon request by the authors of \citet{johnson2019mimic}.}. Details regarding the label distribution of both datasets are attached in Table \ref{tab:datasets} of the Appendix.

\subsection{RadPert Evaluation}

\renewcommand{\arraystretch}{1.3}
\begin{table*}
\centering
{\small
\begin{tabular}{@{\extracolsep{6pt}}lcccc@{}}
\hline
                     & \multicolumn{2}{c}{\textbf{MIMIC-CXR Gold Standard Test Set}}                                                                                          & \multicolumn{2}{c}{\textbf{CUH}}                                                                                                 \\ \cline{2-3} \cline{4-5} 
\textbf{Pathologies} & \textbf{\begin{tabular}[l]{@{}l@{}}Weighted F1\\ RadPert\end{tabular}} & \textbf{\begin{tabular}[l]{@{}l@{}}Improvement over\\ CheXpert (\%)\end{tabular}} & \textbf{\begin{tabular}[l]{@{}l@{}}Weighted F1\\ RadPert\end{tabular}} & \textbf{\begin{tabular}[l]{@{}l@{}}Improvement over\\ CheXpert (\%)\end{tabular}} \\ \hline
Atelectasis          & 0.782{\fontsize{8}{9}\selectfont \phantom{00}(0.740, 0.825)}                                                   &\phantom{0} -5.2{\fontsize{8}{9}\selectfont \phantom{-0}(-10.2, 0.2)}                                                             & 0.893{\fontsize{8}{9}\selectfont \phantom{00}(0.836, 0.941)}                                                   &\phantom{00} -0.8{\fontsize{8}{9}\selectfont \phantom{00-000}(-6.3, 4.4)}                                                              \\
Cardiomegaly         & 0.801{\fontsize{8}{9}\selectfont \phantom{00}(0.749, 0.846)}                                                   &\phantom{-0} 8.1{\fontsize{8}{9}\selectfont \phantom{-0-}(4.2, 12.6)}                                                               & 0.910{\fontsize{8}{9}\selectfont \phantom{00}(0.872, 0.945)}                                                   &\phantom{-0} 27.3{\fontsize{8}{9}\selectfont \phantom{-0-00}(16.4, 41.1)}                                                             \\
Consolidation        & 0.806{\fontsize{8}{9}\selectfont \phantom{00}(0.731, 0.872)}                                                   &\phantom{-} 15.5{\fontsize{8}{9}\selectfont \phantom{-0-}(1.9, 33.4)}                                                              & 0.951{\fontsize{8}{9}\selectfont \phantom{00}(0.928, 0.971)}                                                   &\phantom{-00} 3.0{\fontsize{8}{9}\selectfont \phantom{-00-000}(0.4, 5.8)}                                                                \\
Edema                & 0.801{\fontsize{8}{9}\selectfont \phantom{00}(0.758, 0.843)}                                                   &\phantom{-0} 0.1{\fontsize{8}{9}\selectfont \phantom{0-0}(-5.6, 3.9)}                                                               & 0.625{\fontsize{8}{9}\selectfont \phantom{00}(0.466, 0.754)}                                                   &\phantom{00} -5.5{\fontsize{8}{9}\selectfont \phantom{0-00}(-28.1, 19.1)}                                                            \\
Enlarged Card.       & 0.628{\fontsize{8}{9}\selectfont \phantom{00}(0.548, 0.702)}                                                   &\phantom{-} 23.8{\fontsize{8}{9}\selectfont \phantom{-0-0}(5.6, 4.7)}                                                               & 0.908{\fontsize{8}{9}\selectfont \phantom{00}(0.860, 0.950)}                                                   &\phantom{-00} 0.7{\fontsize{8}{9}\selectfont \phantom{00-000}(-2.1, 3.5)}                                                               \\
Fracture             & 0.866{\fontsize{8}{9}\selectfont \phantom{00}(0.765, 0.946)}                                                   &\phantom{-} 30.8{\fontsize{8}{9}\selectfont \phantom{-0-}(9.7, 60.9)}                                                              & 0.764{\fontsize{8}{9}\selectfont \phantom{00}(0.593, 0.898)}                                                   &\phantom{-0} 12.7{\fontsize{8}{9}\selectfont \phantom{00-00}(-8.1, 47.5)}                                                             \\
Lung Lesion          & 0.696{\fontsize{8}{9}\selectfont \phantom{00}(0.583, 0.797)}                                                   &\phantom{-0} 4.0{\fontsize{8}{9}\selectfont \phantom{0-}(-5.4, 14.8)}                                                              & 0.816{\fontsize{8}{9}\selectfont \phantom{00}(0.706, 0.911)}                                                   &\phantom{-} 660.4{\fontsize{8}{9}\selectfont \phantom{--}(210.8, 2700.3)}                                                         \\
Lung Opacity         & 0.783{\fontsize{8}{9}\selectfont \phantom{00}(0.741, 0.827)}                                                   &\phantom{-0} 3.2{\fontsize{8}{9}\selectfont \phantom{0-0}(-1.3, 8.7)}                                                               & 0.712{\fontsize{8}{9}\selectfont \phantom{00}(0.652, 0.766)}                                                   &\phantom{-00} 0.8{\fontsize{8}{9}\selectfont \phantom{00-000}(-1.6, 3.5)}                                                               \\
Pleur. Effusion      & 0.873{\fontsize{8}{9}\selectfont \phantom{00}(0.843, 0.901)}                                                   &\phantom{0} -3.3{\fontsize{8}{9}\selectfont \phantom{00}(-6.4, -0.2)}                                                             & 0.641{\fontsize{8}{9}\selectfont \phantom{00}(0.587, 0.689)}                                                   &\phantom{-00} 0.1{\fontsize{8}{9}\selectfont \phantom{00-000}(-2.5, 2.8)}                                                               \\
Pleur. Other         & 0.547{\fontsize{8}{9}\selectfont \phantom{00}(0.390, 0.692)}                                                   &\phantom{-} 16.7{\fontsize{8}{9}\selectfont \phantom{-0-}(1.6, 44.0)}                                                              & 0.082{\fontsize{8}{9}\selectfont \phantom{00}(0.043, 0.127)}                                                   &\phantom{-} 189.8{\fontsize{8}{9}\selectfont \phantom{-0-0}(45.9, 713.3)}                                                           \\
Pneumonia            & 0.757{\fontsize{8}{9}\selectfont \phantom{00}(0.704, 0.806)}                                                   &\phantom{-} 28.1{\fontsize{8}{9}\selectfont \phantom{--}(15.8, 42.5)}                                                             & 0.656{\fontsize{8}{9}\selectfont \phantom{00}(0.520, 0.773)}                                                   &\phantom{-0} 54.5{\fontsize{8}{9}\selectfont \phantom{-00-0}(9.4, 130.9)}                                                             \\
Pneumothorax         & 0.898{\fontsize{8}{9}\selectfont \phantom{00}(0.856, 0.934)}                                                   &\phantom{-0} 5.1{\fontsize{8}{9}\selectfont \phantom{0-}(-0.4, 10.9)}                                                              & 0.626{\fontsize{8}{9}\selectfont \phantom{00}(0.568, 0.682)}                                                   &\phantom{-00} 2.1{\fontsize{8}{9}\selectfont \phantom{00-000}(-0.9, 5.7)}                                                               \\
Sup. Devices         & 0.886{\fontsize{8}{9}\selectfont \phantom{00}(0.854, 0.915)}                                                   &\phantom{-0} 2.1{\fontsize{8}{9}\selectfont \phantom{0-0}(-0.4, 5.1)}                                                               & 0.858{\fontsize{8}{9}\selectfont \phantom{00}(0.825, 0.890)}                                                    &\phantom{00} -2.6{\fontsize{8}{9}\selectfont \phantom{00000}(-4.8, -0.6)}                                                             \\ \hline
Macro Avg.           & 0.757{\fontsize{8}{9}\selectfont \phantom{00}(0.779, 0.800)}                                                   &\phantom{-0} 8.0{\fontsize{8}{9}\selectfont \phantom{-0-}(5.5, 10.8)}                                                               & 0.726{\fontsize{8}{9}\selectfont \phantom{00}(0.699, 0.752)}                                                   &\phantom{-0} 14.6{\fontsize{8}{9}\selectfont \phantom{-0-00}(10.4, 19.1)}                                                             \\ 
Weighted Avg.        & 0.816{\fontsize{8}{9}\selectfont \phantom{00}(0.802, 0.830)}                                                   &\phantom{-0} 3.4{\fontsize{8}{9}\selectfont \phantom{-0-0}(1.5, 5.3)}                                                                & 0.787{\fontsize{8}{9}\selectfont \phantom{00}(0.765, 0.808)}                                                   &\phantom{-00} 5.0{\fontsize{8}{9}\selectfont \phantom{-00-000}(2.6, 7.3)}                                                                \\ \hline

\end{tabular}
}
\caption{Weighted average F1 scores for RadPert alongside improvements over the CheXpert model on the MIMIC-CXR gold-standard and CUH test sets. The F1 scores are averaged across the sub-tasks of positive mention detection, negation detection, and uncertainty detection weighted by the support sets. The scores correspond to the averages across 1000 bootstrap replicates and are reported alongside their confidence intervals.}
\label{tab:f1_weight_avg_both}
\end{table*}

In Table \ref{tab:f1_weight_avg_both}, we report the weighted average F1 scores across the sub-tasks of positive mention detection, negation detection, and uncertainty Detection for the MIMIC-CXR and CUH datasets. We are also reporting the improvements over the CheXpert labeler alongside their confidence intervals. Radpert achieves a statistically significant improvement both on average and on the majority of the pathologies. Namely, for MIMIC-CXR, RadPert is 8.0\% (95\% CI: 5.5\%, 10.8\%) better than CheXpert, yielding an average F1 score of 0.757 (95\% CI: 0.779, 0.800). 

In Table \ref{tab:radpert_f1_neg_unc} of the Appendix, we also report fine-grained results in the distinct sub-tasks. In addition to the sub-tasks of negation, positive mention, and uncertainty detection, we also report the performance improvement in mention detection. Mention detection treats \textit{Null} as the positive class, and \textit{Negative}, \textit{Uncertain}, and \textit{Positive} as the negative class.

\subsubsection{Discussion on RadPert's Performance}

We observe performance improvement in all sub-tasks. The strongest improvement is achieved in the uncertainty detection task, showcasing the effectiveness of utilizing the uncertainty labels of RadGraph. However, the improvement in mention detection is marginal. A primary cause of mention detection failure is the reliance on the RadGraph model, which occasionally fails to recall all entities and relations within a radiology report.

Focusing on specific pathologies, RadPert fails to consistently outperform CheXpert for Atelectasis, Edema, and Pleural Effusion. In the case of Atelectasis and Edema, the rule sets are straightforward, and their mentions often lack syntactic variability in practice, offering limited benefit from the uncertainty-aware entity representations of RadGraph. Regarding Pleural Effusion, RadPert is hindered by the divergence between RadGraph annotation guidelines\footnote{Available on \href{https://openreview.net/forum?id=pMWtc5NKd7V}{OpenReview}.} and those of the MIMIC-CXR and CUH datasets concerning uncertainty. Specifically, RadGraph suggests annotating any degree of uncertainty as \textit{OBS:Uncertain} \citep{RadGraphNeurips} while the MIMIC-CXR guidelines, also used by CUH, permit some degree of uncertainty within \textit{Positive} and \textit{Negative} labels. For instance, ``likely representing pneumonia'' should be labeled as positive according to MIMIC-CXR guidelines. For Pleural Effusion, uncertain mentions such as ``minimal if any pleural effusion'' are commonplace and labeled inconsistently by the annotators in MIMIC-CXR. However, due to RadGraph's annotation guidelines, RadPert primarily labels such mentions as \textit{Uncertain}, resulting in low precision in the uncertainty detection task for Pleural Effusion. This behavior can be observed in the Pleural Effusion confusion matrices (Appendix, Figure \ref{fig:conf_mimic}).

Notably, RadPert's performance for Lung Lesion showed a substantial improvement over CheXpert's performance on the CUH dataset compared to MIMIC-CXR. This discrepancy arises because ``lung lesion'' is a specific term frequently used in the CUH reports, while it rarely appears in MIMIC-CXR reports. The CheXpert labeler treats Lung Lesion as an umbrella term encompassing ``masses'', ``nodular opacities'', and ``carcinomata'', lacking specific rules for ``lung lesions'' and only identifying the less general terms, leading to inconsistent performance in CUH. Additionally, variations such as ``edema'' in the US and ``oedema'' in the UK also illustrate the divergent terminology and spelling conventions between the two corpora, although these spelling differences do not affect the ability of CheXpert to detect Edema mentions.

Finally, in Table \ref{tab:co2} of the Appendix, we provide carbon estimates for both CheXpert and RadPert. RadPert not only improves upon CheXpert in performance but also demonstrates greater energy efficiency.

\subsection{RadPrompt Evaluation}

\renewcommand{\arraystretch}{1.3}
\begin{table*}
    \centering
        {\small
            \begin{tabular}{lcccc}
            \hline
                                 & \multicolumn{4}{c}{\textbf{RadPrompt Improvement of Weighted Average F1 Over 1{}\textsuperscript{st} Turn (\%)}} \\ \cline{2-5}
            \textbf{Pathologies} & \textbf{Gemini-1.5 Pro}         & \textbf{Llama-2 70B}         & \textbf{Claude-3 Sonnet}         & \textbf{GPT-4 Turbo}        \\ \hline
            Atelectasis          &\phantom{00} -0.9 {\fontsize{8}{9}\selectfont \phantom{0000}(-4.4, 3.0)}                &\phantom{00} -7.0 {\fontsize{8}{9}\selectfont \phantom{000}(-12.6, -0.2)}           &\phantom{00} -1.4 {\fontsize{8}{9}\selectfont \phantom{0000}(-7.1, 5.3)}                 &\phantom{00} -3.9 {\fontsize{8}{9}\selectfont \phantom{00}(-7.2, -0.4)}           \\
            Cardiomegaly         &\phantom{00} -2.3 {\fontsize{8}{9}\selectfont \phantom{0000}(-6.6, 1.9)}                &\phantom{-0} 14.3 {\fontsize{8}{9}\selectfont \phantom{0-0-0}(9.2, 20.2)}             &\phantom{-00} 2.7 {\fontsize{8}{9}\selectfont \phantom{0000}(-2.4, 7.6)}                  &\phantom{00} -1.9 {\fontsize{8}{9}\selectfont \phantom{0-0}(-5.5, 1.5)}            \\
            Consolidation        &\phantom{-0} 26.6 {\fontsize{8}{9}\selectfont \phantom{0-0}(13.9, 40.5)}               &\phantom{-0} 70.7 {\fontsize{8}{9}\selectfont \phantom{0--}(43.8, 102.6)}           &\phantom{-0} 31.9 {\fontsize{8}{9}\selectfont \phantom{0-0}(15.7, 49.7)}                &\phantom{-00} 2.6 {\fontsize{8}{9}\selectfont \phantom{0-0}(-3.6, 9.4)}             \\
            Edema                &\phantom{-00} 7.7 {\fontsize{8}{9}\selectfont \phantom{00-0}(3.2, 12.5)}                 &\phantom{-0} 10.3 {\fontsize{8}{9}\selectfont \phantom{0-0-0}(4.5, 16.6)}             &\phantom{-00} 7.4 {\fontsize{8}{9}\selectfont \phantom{00-0}(1.9, 13.1)}                  &\phantom{00} -3.1 {\fontsize{8}{9}\selectfont \phantom{00}(-5.9, -0.4)}           \\
            Enlarged Card.       &\phantom{-0} 49.7 {\fontsize{8}{9}\selectfont \phantom{0-0}(22.1, 89.6)}               &\phantom{-} 160.2 {\fontsize{8}{9}\selectfont \phantom{0--}(75.1, 309.4)}          &\phantom{-} 103.0 {\fontsize{8}{9}\selectfont \phantom{0-}(55.7, 167.3)}              &\phantom{-00} 3.9 {\fontsize{8}{9}\selectfont \phantom{0-}(-8.6, 17.3)}            \\
            Fracture             &\phantom{-0} 10.7 {\fontsize{8}{9}\selectfont \phantom{00-0}(1.4, 23.6)}                &\phantom{-0} 20.1 {\fontsize{8}{9}\selectfont \phantom{0-0-0}(4.6, 42.0)}             &\phantom{-0} 14.8 {\fontsize{8}{9}\selectfont \phantom{00-0}(0.8, 31.2)}                 &\phantom{-00} 5.2 {\fontsize{8}{9}\selectfont \phantom{0--0}(0.9, 9.9)}              \\
            Lung Lesion          &\phantom{-0} 65.5 {\fontsize{8}{9}\selectfont \phantom{0-}(37.3, 100.6)}              &\phantom{-0} 24.0 {\fontsize{8}{9}\selectfont \phantom{0-0-0}(3.7, 48.0)}             &\phantom{-00} 3.2 {\fontsize{8}{9}\selectfont \phantom{00}(-11.5, 18.5)}                &\phantom{-00} 6.5 {\fontsize{8}{9}\selectfont \phantom{0-}(-7.0, 20.4)}            \\
            Lung Opacity         &\phantom{-0} 26.9 {\fontsize{8}{9}\selectfont \phantom{0-0}(18.8, 36.2)}               &\phantom{-0} 23.5 {\fontsize{8}{9}\selectfont \phantom{0--0}(15.9, 32.3)}            &\phantom{-0} 23.6 {\fontsize{8}{9}\selectfont \phantom{0-0}(14.1, 34.0)}                &\phantom{-00} 8.1 {\fontsize{8}{9}\selectfont \phantom{0--}(2.2, 14.4)}             \\
            Pleural Effusion     &\phantom{-00} 4.1 {\fontsize{8}{9}\selectfont \phantom{00-00}(1.5, 6.5)}                  &\phantom{-00} 4.9 {\fontsize{8}{9}\selectfont \phantom{0-0-00}(1.2, 9.0)}               &\phantom{-00} 8.3 {\fontsize{8}{9}\selectfont \phantom{00-0}(5.2, 11.4)}                  &\phantom{-00} 0.3 {\fontsize{8}{9}\selectfont \phantom{0-0}(-1.8, 2.4)}             \\
            Pleural Other        &\phantom{-0} 21.0 {\fontsize{8}{9}\selectfont \phantom{00-0}(1.8, 44.6)}                &\phantom{-} 158.3 {\fontsize{8}{9}\selectfont \phantom{00-}(-0.1, 291.8)}          &\phantom{-0} 36.8 {\fontsize{8}{9}\selectfont \phantom{00-0}(8.2, 72.8)}                 &\phantom{-0} 10.8 {\fontsize{8}{9}\selectfont \phantom{0-}(-6.9, 29.4)}           \\
            Pneumonia            &\phantom{-0} 15.6 {\fontsize{8}{9}\selectfont \phantom{0-0}(10.3, 21.4)}               &\phantom{00} -5.3 {\fontsize{8}{9}\selectfont \phantom{0-00}(-14.1, 4.0)}            &\phantom{-0} 22.0 {\fontsize{8}{9}\selectfont \phantom{0-0}(14.2, 30.5)}                &\phantom{-00} 4.5 {\fontsize{8}{9}\selectfont \phantom{0--0}(1.2, 8.3)}              \\
            Pneumothorax         &\phantom{-0} 20.5 {\fontsize{8}{9}\selectfont \phantom{0-0}(14.9, 26.3)}               &\phantom{-0} 19.3 {\fontsize{8}{9}\selectfont \phantom{0--0}(12.7, 26.8)}            &\phantom{-0} 34.9 {\fontsize{8}{9}\selectfont \phantom{0-0}(28.2, 42.5)}                &\phantom{-00} 1.0 {\fontsize{8}{9}\selectfont \phantom{0-0}(-1.3, 3.3)}             \\
            Support Devices      &\phantom{-00} 4.1 {\fontsize{8}{9}\selectfont \phantom{00-00}(1.8, 6.7)}                  &\phantom{-0} 23.1 {\fontsize{8}{9}\selectfont \phantom{0--0}(15.7, 31.7)}            &\phantom{-00} 1.1 {\fontsize{8}{9}\selectfont \phantom{0000}(-0.8, 3.3)}                  &\phantom{-00} 0.5 {\fontsize{8}{9}\selectfont \phantom{0-0}(-0.5, 1.6)}             \\ \hline
            Macro Average        &\phantom{-0} 14.8 {\fontsize{8}{9}\selectfont \phantom{0-0}(12.2, 17.3)}               &\phantom{-0} 20.8 {\fontsize{8}{9}\selectfont \phantom{0--0}(16.2, 25.8)}            &\phantom{-0} 16.2 {\fontsize{8}{9}\selectfont \phantom{0-0}(13.1, 19.4)}                &\phantom{-00} 2.1 {\fontsize{8}{9}\selectfont \phantom{0--0}(0.3, 4.1)}              \\
            Weighted Average     &\phantom{-0} 10.2 {\fontsize{8}{9}\selectfont \phantom{00-0}(8.4, 12.0)}                &\phantom{-0} 12.5 {\fontsize{8}{9}\selectfont \phantom{0-0-0}(9.7, 15.4)}             &\phantom{-0} 12.7 {\fontsize{8}{9}\selectfont \phantom{0-0}(10.7, 15.0)}                &\phantom{-00} 0.9 {\fontsize{8}{9}\selectfont \phantom{0-0}(-0.2, 2.1)}             \\ \hline

            \end{tabular}
            }
    \caption{Improvement of weighted average F1 scores for RadPrompt over the base LLM on MIMIC-CXR gold-standard test set, alongside confidence intervals. Improvement is measured in a multi-turn chat setting by comparing the initial classification decision of the LLM to the revised classification decision after introducing RadPert hints.}
    \label{tab:radprompt_improve}
    \vspace{12pt}
            {\small
        \begin{tabular}{lcccc}
                             \hline
                             & \multicolumn{4}{c}{\textbf{RadPrompt Improvement of Weighted Average F1 Over RadPert (\%)}}      \\ \cline{2-5}
        \textbf{Pathologies} & \textbf{Gemini-1.5 Pro} & \textbf{Llama-2 70B} & \textbf{Claude-3 Sonnet} & \textbf{GPT-4 Turbo} \\ \hline
        Atelectasis          &\phantom{-0} 6.2 {\fontsize{8}{9}\selectfont \phantom{0--}(0.8, 11.7)}         &\phantom{0} -0.0 {\fontsize{8}{9}\selectfont \phantom{00-0}(-1.6, 1.7)}     &\phantom{-0} 3.8 {\fontsize{8}{9}\selectfont \phantom{0-0-0}(0.6, 7.5)}           &\phantom{-0} 6.2 {\fontsize{8}{9}\selectfont \phantom{0-0-}(0.7, 11.7)}      \\
        Cardiomegaly         &\phantom{0} -1.4 {\fontsize{8}{9}\selectfont \phantom{00-}(-4.0, 1.2)}        &\phantom{-0} 0.7 {\fontsize{8}{9}\selectfont \phantom{00-0}(-0.9, 2.4)}      &\phantom{0} -0.2 {\fontsize{8}{9}\selectfont \phantom{00-0}(-1.5, 1.0)}         &\phantom{-0} 0.7 {\fontsize{8}{9}\selectfont \phantom{00-0}(-2.8, 4.5)}      \\
        Consolidation        &\phantom{0} -7.7 {\fontsize{8}{9}\selectfont \phantom{0-}(-16.0, 0.1)}       &\phantom{} -22.4 {\fontsize{8}{9}\selectfont \phantom{0}(-29.6, -16.2)} &\phantom{0} -0.6 {\fontsize{8}{9}\selectfont \phantom{00-0}(-4.2, 3.2)}         &\phantom{-0} 2.4 {\fontsize{8}{9}\selectfont \phantom{00-0}(-3.8, 9.1)}      \\
        Edema                &\phantom{0} -0.9 {\fontsize{8}{9}\selectfont \phantom{00-}(-3.9, 2.3)}        &\phantom{-0} 0.5 {\fontsize{8}{9}\selectfont \phantom{00-0}(-0.8, 1.9)}      &\phantom{-0} 0.1 {\fontsize{8}{9}\selectfont \phantom{00-0}(-0.8, 1.2)}          &\phantom{-0} 1.3 {\fontsize{8}{9}\selectfont \phantom{00-0}(-1.7, 4.7)}      \\
        Enlarged Card.       &\phantom{} -11.6 {\fontsize{8}{9}\selectfont \phantom{0}(-19.1, -5.1)}     &\phantom{0} -1.5 {\fontsize{8}{9}\selectfont \phantom{00-0}(-4.0, 0.7)}     &\phantom{0} -8.0 {\fontsize{8}{9}\selectfont \phantom{00}(-14.1, -2.4)}       &\phantom{0} -6.5 {\fontsize{8}{9}\selectfont \phantom{00}(-12.8, -0.8)}   \\
        Fracture             &\phantom{0} -8.5 {\fontsize{8}{9}\selectfont \phantom{0}(-15.7, -1.2)}      &\phantom{0} -1.2 {\fontsize{8}{9}\selectfont \phantom{00-0}(-4.0, 1.2)}     &\phantom{0} -2.0 {\fontsize{8}{9}\selectfont \phantom{00-0}(-5.2, 0.0)}         &\phantom{0} -4.5 {\fontsize{8}{9}\selectfont \phantom{0-0}(-11.7, 3.3)}    \\
        Lung Lesion          &\phantom{0} -2.4 {\fontsize{8}{9}\selectfont \phantom{00-}(-9.2, 4.9)}        &\phantom{} -28.4 {\fontsize{8}{9}\selectfont \phantom{0}(-37.9, -19.4)} &\phantom{-0} 2.1 {\fontsize{8}{9}\selectfont \phantom{00-}(-5.3, 11.1)}         &\phantom{0} -2.9 {\fontsize{8}{9}\selectfont \phantom{0-0}(-14.0, 9.2)}    \\
        Lung Opacity         &\phantom{0} -5.0 {\fontsize{8}{9}\selectfont \phantom{00}(-8.0, -2.1)}       &\phantom{0} -0.4 {\fontsize{8}{9}\selectfont \phantom{00-0}(-1.9, 1.1)}     &\phantom{0} -0.4 {\fontsize{8}{9}\selectfont \phantom{00-0}(-2.7, 1.8)}         &\phantom{0} -0.2 {\fontsize{8}{9}\selectfont \phantom{00-0}(-3.1, 2.8)}     \\
        Pleural Effusion     &\phantom{-0} 2.0 {\fontsize{8}{9}\selectfont \phantom{0-0-}(0.0, 4.1)}          &\phantom{0} -0.7 {\fontsize{8}{9}\selectfont \phantom{00-0}(-2.1, 0.9)}     &\phantom{-0} 3.2 {\fontsize{8}{9}\selectfont \phantom{0-0-0}(1.6, 5.0)}           &\phantom{-0} 2.8 {\fontsize{8}{9}\selectfont \phantom{0-0-0}(0.4, 5.4)}       \\
        Pleural Other        &\phantom{} -10.0 {\fontsize{8}{9}\selectfont \phantom{0-}(-20.3, 1.8)}      &\phantom{0} -4.0 {\fontsize{8}{9}\selectfont \phantom{0-0}(-12.5, 0.0)}    &\phantom{-0} 0.0 {\fontsize{8}{9}\selectfont \phantom{0-0-0}(0.0, 0.0)}           &\phantom{-} 13.5 {\fontsize{8}{9}\selectfont \phantom{00-}(-3.9, 39.7)}    \\
        Pneumonia            &\phantom{-0} 4.2 {\fontsize{8}{9}\selectfont \phantom{00-}(-0.1, 9.4)}         &\phantom{} -14.8 {\fontsize{8}{9}\selectfont \phantom{00}(-19.8, -9.7)}  &\phantom{-0} 3.0 {\fontsize{8}{9}\selectfont \phantom{0-0-0}(0.5, 6.4)}           &\phantom{-0} 4.4 {\fontsize{8}{9}\selectfont \phantom{00-0}(-0.4, 9.5)}      \\
        Pneumothorax         &\phantom{0} -0.6 {\fontsize{8}{9}\selectfont \phantom{00-}(-3.1, 2.1)}        &\phantom{0} -3.0 {\fontsize{8}{9}\selectfont \phantom{000}(-5.0, -1.3)}    &\phantom{-0} 2.7 {\fontsize{8}{9}\selectfont \phantom{0-0-0}(0.3, 5.6)}           &\phantom{-0} 3.5 {\fontsize{8}{9}\selectfont \phantom{0-0-0}(0.8, 7.1)}       \\
        Support Devices      &\phantom{-0} 2.2 {\fontsize{8}{9}\selectfont \phantom{0-0-}(0.5, 4.0)}          &\phantom{0} -0.2 {\fontsize{8}{9}\selectfont \phantom{00-0}(-1.2, 0.5)}     &\phantom{-0} 1.2 {\fontsize{8}{9}\selectfont \phantom{00-0}(-0.0, 2.5)}          &\phantom{-0} 0.2 {\fontsize{8}{9}\selectfont \phantom{00-0}(-2.4, 2.8)}      \\ \hline
        Macro Average        &\phantom{0} -2.2 {\fontsize{8}{9}\selectfont \phantom{00}(-3.8, -0.6)}       &\phantom{0} -5.5 {\fontsize{8}{9}\selectfont \phantom{000}(-6.9, -4.3)}    &\phantom{-0} 0.5 {\fontsize{8}{9}\selectfont \phantom{00-0}(-0.4, 1.4)}          &\phantom{-0} 1.4 {\fontsize{8}{9}\selectfont \phantom{00-0}(-0.5, 3.2)}      \\
        Weighted Average     &\phantom{0} -0.2 {\fontsize{8}{9}\selectfont \phantom{00-}(-1.5, 1.2)}        &\phantom{0} -3.5 {\fontsize{8}{9}\selectfont \phantom{000}(-4.4, -2.7)}    &\phantom{-0} 1.4 {\fontsize{8}{9}\selectfont \phantom{0-0-0}(0.7, 2.1)}           &\phantom{-0} 1.9 {\fontsize{8}{9}\selectfont \phantom{0-0-0}(0.7, 3.2)}       \\ \hline
        \end{tabular}
        }
            \caption{Improvement of weighted average F1 scores for RadPrompt over the rule-based RadPert on the MIMIC-CXR gold-standard test set, alongside confidence intervals.}
    \label{tab:radprompt_improve_radpert}
\end{table*}

In Table \ref{tab:radprompt_improve}, we present the improvement in the weighted average F1 score of RadPrompt for various base LLMs on the MIMIC-CXR gold-standard test set. Specifically, we compare the revised classification outcome of the second-turn prompt, which is infused with RadPert hints, to the first-turn classification outcome. For all tested LLMs, we observe that the RadPrompt strategy leads, on average (across pathologies), to a statistically significant improvement over the baseline zero-shot prompting. For clarity, in Tables \ref{tab:radprompt_f1}, \ref{tab:radprompt_f1_mention}, \ref{tab:radprompt_f1_negation}, \ref{tab:radprompt_f1_uncertainty}
and \ref{tab:radprompt_f1_positive} of the Appendix, we also report the task-specific F1 scores of the first and second turns of RadPrompt.

Furthermore, we compare RadPrompt's second-turn results with RadPert in Table \ref{tab:radprompt_improve_radpert} for the MIMIC-CXR gold-standard test set. On average, RadPrompt with Gemini-1.5 Pro and Llama-2 70 B fail to outperform RadPert. However, Claude-3 Sonnet and GPT-4 Turbo-based RadPrompt surpass RadPert. 

Regarding the external evaluation of RadPrompt, the current ethical agreement with the Cambridge University Hospitals limits the use of third-party APIs. Thus, we are only able to evaluate RadPrompt with a Llama-2 base. We present the weighted average and the sub-task-specific results in Tables \ref{tab:radprompt_llama2_cuh} and \ref{tab:llama_cuh_fine}. Similarly to the MIMIC-CXR gold-standard test set, we observe that Llama-2-based RadPrompt enhances the performance of Llama-2 but fails to improve upon RadPert.



\subsubsection{Discussion on RadPrompt's Performance}

\renewcommand{\arraystretch}{1.3}
\begin{table*}
\centering
{\small
    \begin{tabular}{lcccc}
                           \hline
                           & \multicolumn{4}{c}{\textbf{RadPrompt Improvement of Weighted Average F1 Over 1\textsuperscript{st} Turn (\%)}}     \\ \cline{2-5}
    \textbf{Sub-task}      & \textbf{Gemini-1.5 Pro} & \textbf{Llama-2 70B} & \textbf{Claude-3 Sonnet} & \textbf{GPT-4 Turbo} \\ \hline
    Mention Detection      &\phantom{} 17.8{\fontsize{8}{9}\selectfont \phantom{0-}(15.6, 20.0)}       &\phantom{-} 26.7{\fontsize{8}{9}\selectfont \phantom{0-}(23.8, 29.8)}    &\phantom{-} 24.6{\fontsize{8}{9}\selectfont \phantom{0-}(21.7, 27.8)}        &\phantom{-} 1.9{\fontsize{8}{9}\selectfont \phantom{0-0}(0.9, 3.0)}       \\ \hline
    Negation Detection     &\phantom{} 31.9{\fontsize{8}{9}\selectfont \phantom{0-}(26.4, 37.6)}       &\phantom{-} 54.8{\fontsize{8}{9}\selectfont \phantom{0-}(45.8, 64.2)}    &\phantom{-} 62.3{\fontsize{8}{9}\selectfont \phantom{0-}(52.1, 73.1)}        &\phantom{-} 4.9{\fontsize{8}{9}\selectfont \phantom{0-0}(2.3, 8.1)}       \\
    Pos. Mention Detection &\phantom{0} 3.8{\fontsize{8}{9}\selectfont \phantom{00-0}(2.4, 5.2)}          &\phantom{-0} 1.7{\fontsize{8}{9}\selectfont \phantom{000}(-0.5, 4.1)}      &\phantom{-0} 0.7{\fontsize{8}{9}\selectfont \phantom{000}(-0.9, 2.4)}          &\phantom{} -0.4{\fontsize{8}{9}\selectfont \phantom{00}(-1.6, 0.7)}     \\
    Uncertainty Detection  &\phantom{0} 2.9{\fontsize{8}{9}\selectfont \phantom{00}(-5.9, 13.0)}        &\phantom{0} -6.4{\fontsize{8}{9}\selectfont \phantom{00}(-20.7, 9.8)}    &\phantom{0} -0.5{\fontsize{8}{9}\selectfont \phantom{0}(-13.0, 14.0)}       &\phantom{} -2.6{\fontsize{8}{9}\selectfont \phantom{0}(-10.3, 5.9)}    \\ \hline
    Weighted Average       &\phantom{} 10.2{\fontsize{8}{9}\selectfont \phantom{00-}(8.4, 12.0)}        &\phantom{-} 12.5{\fontsize{8}{9}\selectfont \phantom{00-}(9.7, 15.4)}     &\phantom{-} 12.7{\fontsize{8}{9}\selectfont \phantom{0-}(10.7, 15.0)}        &\phantom{-} 0.9{\fontsize{8}{9}\selectfont \phantom{00}(-0.2, 2.1)}      \\ \hline
    \end{tabular}
    }
    \caption{Improvement of RadPrompt over the base LLM for the different sub-tasks on MIMIC-CXR gold-standard test set. For each sub-task. we report the improvement of the weighted average F1 score across all pathologies, along with confidence intervals. The weighted average refers to averaging over sub-tasks, excluding the mention detection task.}
\label{tab:radprompt_err_analysis}
\end{table*}

We can observe from Tables \ref{tab:radprompt_improve} and \ref{tab:radprompt_improve_radpert} that RadPrompt on Claude-3 Sonnet and on GPT-4 Turbo exceeds, on average, both RadPert and the initial LLM predictions. Namely, RadPrompt with GPT-4 Turbo is 2.1\% (CI 0.3\%, 4.1\%) better than baseline GPT-4 Turbo and 1.4\% (CI -0.5\%, 3.2\%) better than RadPert.

Focusing on individual pathologies, we notice that RadPrompt with a Gemini-1.5 Pro base manages to outperform both of its underlying models for Pleural Effusion, Pneumonia, and Support Devices. Additionally, RadPrompt with Claude-3 Sonnet surpasses its underlying models in the case of Lung Lesion, Pleural Effusion, Pneumonia, Pneumothorax, and Support Devices. For a GPT-4 Turbo base, the same behavior is observed for Consolidation, Pleural Effusion, Pleural Other, Pneumonia, and Pneumothorax. The ability of RadPrompt to boost the performance of both its underlying models demonstrates the potential of combining the language reasoning capabilities of LLMs with the insights encoded in rule-based models.

In Table \ref{tab:radprompt_err_analysis}, we present a fine-grained comparison between the first and second turns of RadPrompt. We observe that all models, with the exception of GPT-4 Turbo, initially struggled to understand that we intended to classify only those pathologies explicitly mentioned in the report. This effect disproportionately affects the \textit{Negative} class since \textit{Null} is often conflated with \textit{Negative}. The distinction, however, between those two labels is non-negligible. Inconsistencies often exist between the gold-standard labels extracted directly from chest X-ray Images and the gold-standard labels of their corresponding radiology reports, and thus, pathologies visible within a chest X-ray may be excluded from the radiology report \citep{visualchexbert}. Such observations are also noted in other clinical domains, such as Magnetic Resonance Imaging (MRI), where the clinical context and the referrer physician may bias the observations mentioned within a radiology report \citep{wood2020labelling}.



\section{Limitations}


While this study demonstrates promising improvements in radiology report classification using the RadPrompt methodology, several limitations must be considered.

RadPert and RadPrompt are exclusively developed and tested for the English language. The study also centers around a list of pathologies typical of chest X-rays. As such, the extension of our methodologies to other languages, types of medical imaging, and additional pathologies was not verified.

Furthermore, previous studies have highlighted discrepancies between labels from radiology report annotations and those from the corresponding imaging study annotations \citep{visualchexbert, wood2020labelling}. The source of such inconsistencies includes incomplete radiology report impressions, hierarchical relationships within labels, and the undeniable uncertainty of the task. In future work, we aim to study this effect within the CUH test set.

Due to ethical considerations, we are currently unable to perform inference for the CUH test set through third-party APIs. Thus, we have not evaluated RadPrompt externally for SOTA LLMs. We expect to overcome this limitation after the planned release of the CUH dataset.

Additionally, we cannot estimate the computational cost and carbon footprint for GPT-4-based RadPrompt due to a lack of specific metrics. In the Appendix, we provide carbon footprint estimates for the Llama-2-based RadPrompt, which is significantly higher than RadPert and CheXpert. Nonetheless, RadPert delivers performance comparable to GPT-4 while operating on a commercial CPU with minimal carbon emissions, underscoring its benefits in resource-limited environments.

Finally, there is an inherent degree of ambiguity in classifying radiology reports, especially as it pertains to the \textit{Uncertainty} labels. We aim to extend current datasets with labels from multiple annotators in order to measure annotator agreement.

\section{Conclusions}

This paper introduced RadPert, a rule-based system enhanced by the RadGraph information schema, demonstrating significant improvements in the classification of radiology reports. By leveraging entity-level uncertainty labels, RadPert reduces reliance on comprehensive rule sets. Our evaluations show that RadPert surpasses CheXpert, the previous rule-based SOTA, by achieving an 8.0\% (95\% CI: 5.5\%, 10.8\%) increase in F1 score, with confidence intervals strongly supporting this improvement.

Further extending the application of RadPert, we developed RadPrompt, a multi-turn prompting strategy that utilizes insights from RadPert to enhance the zero-shot prediction capabilities of large language models. RadPrompt demonstrated a 2.1\% (95\% CI: 0.3\%, 4.1\%) improvement in F1 score over GPT-4 Turbo, indicating its potential to refine predictions in clinical settings. These results highlight the growing synergy between structured rule-based systems and large language models, offering a promising direction for future research in biomedical Natural Language Processing. 

As we continue to refine these tools, future work will focus on expanding the existing datasets and addressing the discrepancies between gold-standard image labels and those extracted from radiology reports.

\section*{Code and Data Availability}

Code for RadPert and RadPrompt is available on GitHub\footnote{\href{https://github.com/PanagiotisFytas/RadPert-RadPrompt}{https://github.com/PanagiotisFytas/RadPert-RadPrompt}.}. The CUH dataset is planned to be released in the following months while managed and made available through the hospital's clinical informatics unit.

\section*{Ethical Considerations}

For the MIMIC-CXR gold-standard test set, access to LLMs through APIs conforms to the PhysioNet responsible use guidelines\footnote{\href{https://physionet.org/news/post/gpt-responsible-use}{https://physionet.org/news/post/gpt-responsible-use}}.

This ethical agreement with Cambridge University Hospitals currently limits the use of third-party APIs, but it is being revised prior to the dataset's release.

\section*{Acknowledgments}

Panagiotis wishes to thank the David and Claudia Harding Foundation for their support through the University of Cambridge Harding Distinguished Postgraduate Scholars Programme. Anna Breger and Ian wish to acknowledge the EU/EFPIA Innovative Medicines Initiative 2 Joint Undertaking - DRAGON (101005122). Anna Breger is also supported by the Austrian Science Fund (FWF) through project P33217 and Ian by the National Institute for Health and Care Research (NIHR) Cambridge Biomedical Research Centre (BRC-1215-20014). Anna Korhonen wishes to acknowledge the UK Research and Innovation (UKRI) Frontier Research Grant (EP/Y031350/1).

 \newpage

\bibliography{acl_latex}

\appendix

\section*{Appendix}
\label{sec:appendix}



\begin{table*}
\centering
{\small
\begin{tabular}{lllll}
\hline
                       & \textbf{CheXpert}        & \textbf{RadPert}         & \textbf{Llama-2 70B} & \textbf{RadPrompt /w Llama-2 70B} \\ \hline
\textbf{Runtime (min)} & 7.1                      & 4.8                      & 41.8                 & 43.6                              \\
\textbf{CO\textsubscript{2}e (g)}      & 5.44                     & 3.68                     & 85.48                & 89.16                             \\
\textbf{Device}        & CPU                      & CPU                      & GPU                  & GPU                               \\
\textbf{Model}         & Core i7-6700k & Core i7-6700k & NVIDIA RTX 6000 Ada  & NVIDIA RTX 6000 Ada               \\ \hline
\end{tabular}
}
\caption{Carbon footprint for inference on both MIMIC-CXR gold-standard and CUH test sets, as estimated utilizing the tools from \citet{lannelongue2021green}. For RadPert, calculations include the extraction of the RadGraph knowledge graph. Notably, we are not able to provide estimates for GPT-4 Turbo, Gemini-1.5 Pro, and Claude-3 Sonnet since this information is not provided by the respective API providers.}
\label{tab:co2}
\end{table*}


\renewcommand{\arraystretch}{1.3}
\begin{table*}
\centering
{\small
\begin{tabular}{@{\extracolsep{6pt}}lcccc@{}}
\hline
\multicolumn{1}{c}{\textbf{}} & \multicolumn{2}{c}{\textbf{Negation Detection}}                                                 & \multicolumn{2}{c}{\textbf{Uncertainty Detection}}                                              \\ \cline{2-3} \cline{4-5}
\textbf{Pathologies}          & \textbf{\begin{tabular}[c]{@{}l@{}}F1 Score\\ RadPert\end{tabular}} & \textbf{\begin{tabular}[c]{@{}l@{}}Improvement over\\ CheXpert (\%)\end{tabular}} & \textbf{\begin{tabular}[c]{@{}l@{}}F1 Score\\ RadPert\end{tabular}} & \textbf{\begin{tabular}[c]{@{}l@{}}Improvement over\\ CheXpert (\%)\end{tabular}} \\ \hline
Atelectasis                   & 0.581{\fontsize{8}{9}\selectfont \phantom{00}(0.000, 0.909)}   &\phantom{-0} 61.6 \phantom{0-}(-41.8, 340.2)                                                               & 0.386{\fontsize{8}{9}\selectfont \phantom{00}(0.256, 0.511)} &\phantom{-00} 0.1 \phantom{-0}(-29.1, 44.7)                                                                 \\
Cardiomegaly                  & 0.834{\fontsize{8}{9}\selectfont \phantom{00}(0.769, 0.892)} &\phantom{-00} 7.1 \phantom{-00-0}(0.6, 14.8)                                                                   & 0.093{\fontsize{8}{9}\selectfont \phantom{00}(0.000, 0.227)}   &\phantom{-0} Inf. \phantom{-0-0}(0.0, Inf.)                                                                   \\
Consolidation                 & 0.877{\fontsize{8}{9}\selectfont \phantom{00}(0.762, 0.960)}  &\phantom{00} -6.2 \phantom{0-00}(-17.4, 2.8)                                                                 & 0.665{\fontsize{8}{9}\selectfont \phantom{00}(0.488, 0.818)} &\phantom{-} 269.7 \phantom{-0-}(0.0, 909.7)                                                                \\
Edema                         & 0.832{\fontsize{8}{9}\selectfont \phantom{00}(0.773, 0.886)} &\phantom{-00} 8.7 \phantom{-00-0}(2.6, 16.5)                                                                   & 0.395{\fontsize{8}{9}\selectfont \phantom{00}(0.160, 0.600)}    &\phantom{-} 104.2 \phantom{-0-}(3.4, 275.3)                                                                \\
Enlarged Card.                & 0.916{\fontsize{8}{9}\selectfont \phantom{00}(0.836, 0.982)} &\phantom{-0} 49.5 \phantom{-0-0}(21.9, 96.6)                                                                 & 0.062{\fontsize{8}{9}\selectfont \phantom{00}(0.000, 0.207)}   &\phantom{00} -3.3 \phantom{-0}(-28.6, 23.1)                                                                \\
Fracture                      & 0.733{\fontsize{8}{9}\selectfont \phantom{00}(0.444, 0.947)} &\phantom{-00} 0.0 \phantom{-00-00}(0.0, 0.0)                                                                    & 0.498{\fontsize{8}{9}\selectfont \phantom{00}(0.000, 1.000)}     &\phantom{-0} Inf. \phantom{-0-0}(0.0, Inf.)                                                                   \\
Lung Lesion                   & 0.422{\fontsize{8}{9}\selectfont \phantom{00}(0.000, 0.800)}     &\phantom{00} -5.1 \phantom{0-0}(-50.0, 55.6)                                                                & 0.128{\fontsize{8}{9}\selectfont \phantom{00}(0.000, 0.400)}     &\phantom{-0} Inf. \phantom{-0-0}(0.0, Inf.)                                                                   \\
Lung Opacity                  & 0.513{\fontsize{8}{9}\selectfont \phantom{00}(0.353, 0.674)} &\phantom{-0} 32.2 \phantom{0-}(-17.3, 128.6)                                                               & 0.000{\fontsize{8}{9}\selectfont \phantom{00}(0.000, 0.000)}       &\phantom{-00} 0.0 \phantom{-0-00}(0.0, 0.0)                                                                    \\
Pler. Effusion              & 0.916{\fontsize{8}{9}\selectfont \phantom{00}(0.871, 0.956)} &\phantom{00} -2.6 \phantom{00-00}(-6.3, 1.3)                                                                  & 0.422{\fontsize{8}{9}\selectfont \phantom{00}(0.267, 0.561)} &\phantom{0} -14.5 \phantom{-0}(-42.6, 22.8)                                                               \\
Pler. Other                 & 0.000{\fontsize{8}{9}\selectfont \phantom{00}(0.000, 0.000)}       &\phantom{-00} 0.0 \phantom{-00-00}(0.0, 0.0)                                                                    & 0.000{\fontsize{8}{9}\selectfont \phantom{00}(0.000, 0.000)}       &\phantom{-00} 0.0 \phantom{-0-00}(0.0, 0.0)                                                                    \\
Pneumonia                     & 0.915{\fontsize{8}{9}\selectfont \phantom{00}(0.867, 0.955)} &\phantom{-0} 17.3 \phantom{-00-0}(8.6, 29.0)                                                                  & 0.671{\fontsize{8}{9}\selectfont \phantom{00}(0.582, 0.743)} &\phantom{-0} 43.8 \phantom{--0}(19.1, 76.5)                                                                 \\
Pneumothorax                  & 0.937{\fontsize{8}{9}\selectfont \phantom{00}(0.912, 0.960)}  &\phantom{-00} 2.1 \phantom{00-00}(-0.7, 5.2)                                                                   & 0.645{\fontsize{8}{9}\selectfont \phantom{00}(0.307, 0.909)} &\phantom{-} 125.6 \phantom{0-}(-7.7, 540.1)                                                               \\
Sup. Devices               & 0.283{\fontsize{8}{9}\selectfont \phantom{00}(0.000, 0.545)}   &\phantom{-00} 0.0 \phantom{-00-00}(0.0, 0.0)                                                                    & 0.000{\fontsize{8}{9}\selectfont \phantom{00}(0.000, 0.000)}       &\phantom{-00} 0.0 \phantom{-0-00}(0.0, 0.0)                                                                    \\ \hline
Macro Avg.                 & 0.743{\fontsize{8}{9}\selectfont \phantom{00}(0.686, 0.810)}  &\phantom{-00} 4.1 \phantom{00-0}(-4.9, 14.5)                                                                  & 0.453{\fontsize{8}{9}\selectfont \phantom{00}(0.369, 0.554)} &\phantom{-0} 40.4 \phantom{-0-0}(9.5, 79.8)                                                                  \\
Weighted Avg.              & 0.872{\fontsize{8}{9}\selectfont \phantom{00}(0.852, 0.893)} &\phantom{-00} 5.9 \phantom{-00-00}(3.3, 8.7)                                                                    & 0.530{\fontsize{8}{9}\selectfont \phantom{00}(0.460, 0.607)}   &\phantom{-0} 31.4 \phantom{-0-0}(8.2, 61.3)                                                                  \\ \hline
\multicolumn{1}{c}{\textbf{}} & \multicolumn{2}{c}{\textbf{Positive Mention Detection}}                                         & \multicolumn{2}{c}{\textbf{Mention Detection}}                                                  \\ \cline{2-3} \cline{4-5}
\textbf{Pathologies}          & \textbf{\begin{tabular}[c]{@{}l@{}}F1 Score\\ RadPert\end{tabular}} & \textbf{\begin{tabular}[c]{@{}l@{}}Improvement over\\ CheXpert (\%)\end{tabular}} & \textbf{\begin{tabular}[c]{@{}l@{}}F1 Score\\ RadPert\end{tabular}} & \textbf{\begin{tabular}[c]{@{}l@{}}Improvement over\\ CheXpert (\%)\end{tabular}} \\ \hline
Atelectasis                   & 0.819{\fontsize{8}{9}\selectfont \phantom{00}(0.776, 0.859)} &\phantom{00} -5.8 \phantom{000}(-10.2, -1.5)                                                                & 0.944{\fontsize{8}{9}\selectfont \phantom{00}(0.920, 0.965)} &\phantom{-00} 0.0 \phantom{-0-00}(0.0, 0.0)                                                                    \\
Cardiomegaly                  & 0.851{\fontsize{8}{9}\selectfont \phantom{00}(0.806, 0.893)} &\phantom{-00} 7.5 \phantom{-00-0}(3.4, 12.0)                                                                   & 0.858{\fontsize{8}{9}\selectfont \phantom{00}(0.826, 0.890)} &\phantom{00} -0.0 \phantom{0-00}(-2.8, 3.0)                                                                  \\
Consolidation                 & 0.815{\fontsize{8}{9}\selectfont \phantom{00}(0.724, 0.885)} &\phantom{-00} 8.6 \phantom{00-0}(-0.6, 19.8)                                                                  & 0.930{\fontsize{8}{9}\selectfont \phantom{00}(0.888, 0.963)} &\phantom{-00} 0.0 \phantom{-0-00}(0.0, 0.0)                                                                    \\
Edema                         & 0.809{\fontsize{8}{9}\selectfont \phantom{00}(0.759, 0.859)} &\phantom{00} -8.2 \phantom{000}(-12.9, -3.3)                                                                & 0.887{\fontsize{8}{9}\selectfont \phantom{00}(0.859, 0.916)} &\phantom{00} -0.2 \phantom{0-00}(-0.9, 0.4)                                                                  \\
Enlarged Card.                & 0.442{\fontsize{8}{9}\selectfont \phantom{00}(0.336, 0.551)} &\phantom{-00} 1.3 \phantom{0-0}(-21.3, 28.8)                                                                 & 0.529{\fontsize{8}{9}\selectfont \phantom{00}(0.454, 0.609)} &\phantom{-0} 16.1 \phantom{0-0}(-0.7, 36.7)                                                                 \\
Fracture                      & 0.902{\fontsize{8}{9}\selectfont \phantom{00}(0.831, 0.964)} &\phantom{-00} 8.1 \phantom{-00-0}(0.9, 17.5)                                                                   & 0.952{\fontsize{8}{9}\selectfont \phantom{00}(0.907, 0.990)} &\phantom{-00} 5.4 \phantom{0-0}(-0.1, 12.4)                                                                  \\
Lung Lesion                   & 0.796{\fontsize{8}{9}\selectfont \phantom{00}(0.702, 0.878)} &\phantom{-00} 1.9 \phantom{00-0}(-6.4, 10.2)                                                                  & 0.834{\fontsize{8}{9}\selectfont \phantom{00}(0.752, 0.901)} &\phantom{00} -2.4 \phantom{0-00}(-7.0, 1.7)                                                                  \\
Lung Opacity                  & 0.819{\fontsize{8}{9}\selectfont \phantom{00}(0.774, 0.859)} &\phantom{-00} 1.6 \phantom{00-00}(-1.3, 4.5)                                                                   & 0.800{\fontsize{8}{9}\selectfont \phantom{00}(0.757, 0.840)} &\phantom{00} -0.0 \phantom{0-00}(-1.2, 1.2)                                                                  \\
Pler. Effusion              & 0.889{\fontsize{8}{9}\selectfont \phantom{00}(0.859, 0.916)} &\phantom{00} -3.1 \phantom{00-00}(-6.3, 0.0)                                                                  & 0.979{\fontsize{8}{9}\selectfont \phantom{00}(0.968, 0.989)} &\phantom{-00} 0.6 \phantom{0-00}(-0.1, 1.4)                                                                   \\
Pler. Other                 & 0.592{\fontsize{8}{9}\selectfont \phantom{00}(0.441, 0.727)} &\phantom{-0} 16.7 \phantom{-00-0}(1.6, 44.0)                                                                  & 0.592{\fontsize{8}{9}\selectfont \phantom{00}(0.459, 0.709)} &\phantom{-00} 1.1 \phantom{0-0}(-5.3, 11.3)                                                                  \\
Pneumonia                     & 0.654{\fontsize{8}{9}\selectfont \phantom{00}(0.550, 0.744)}  &\phantom{-0} 36.9 \phantom{-00-0}(8.5, 75.6)                                                                  & 0.952{\fontsize{8}{9}\selectfont \phantom{00}(0.931, 0.971)} &\phantom{00} -0.5 \phantom{0-00}(-1.5, 0.4)                                                                  \\
Pneumothorax                  & 0.765{\fontsize{8}{9}\selectfont \phantom{00}(0.630, 0.870)}   &\phantom{-00} 9.6 \phantom{00-0}(-7.5, 30.4)                                                                  & 0.963{\fontsize{8}{9}\selectfont \phantom{00}(0.945, 0.980)} &\phantom{00} -0.7 \phantom{0-00}(-1.5, 0.0)                                                                  \\
Sup. Devices               & 0.898{\fontsize{8}{9}\selectfont \phantom{00}(0.869, 0.926)} &\phantom{-00} 1.3 \phantom{00-00}(-0.7, 3.5)                                                                   & 0.893{\fontsize{8}{9}\selectfont \phantom{00}(0.862, 0.918)} &\phantom{-00} 1.4 \phantom{0-00}(-0.2, 3.1)                                                                   \\ \hline
Macro Avg.                 & 0.773{\fontsize{8}{9}\selectfont \phantom{00}(0.749, 0.796)} &\phantom{-00} 4.1 \phantom{-00-00}(1.5, 6.8)                                                                    & 0.855{\fontsize{8}{9}\selectfont \phantom{00}(0.839, 0.870)} &\phantom{-00} 1.0 \phantom{0-00}(-0.0, 2.0)                                                                   \\
Weighted Avg.              & 0.824{\fontsize{8}{9}\selectfont \phantom{00}(0.809, 0.839)} &\phantom{-00} 0.9 \phantom{00-00}(-0.8, 2.6)                                                                   & 0.899{\fontsize{8}{9}\selectfont \phantom{00}(0.890, 0.908)} &\phantom{-00} 0.4 \phantom{0-00}(-0.1, 0.9)                                                                   \\ \hline

\end{tabular}
}
\caption{F1 scores of RadPert and improvement over CheXpert on MIMIC-CXR gold-standard test set. We report results for the sub-tasks of negation detection, uncertainty detection, positive mention detection and mention detection. The scores correspond to the averages across 1000 bootstrap replicates and are reported alongside their confidence intervals.}
\label{tab:radpert_f1_neg_unc}
\end{table*}


\renewcommand{\arraystretch}{1.3}
\begin{table*}
\centering
{\small
\begin{tabular}{@{\extracolsep{6pt}}lcccc@{}}
                     \hline
                     & \multicolumn{4}{c}{\textbf{Weighted Average F1 Across Tasks}}                                                                 \\ \cline{2-5}
                     & \multicolumn{2}{c}{\textbf{Gemini-1.5 Pro}}                   & \multicolumn{2}{c}{\textbf{Llama-2 70B}}                      \\ \cline{2-3} \cline{4-5}
\textbf{Pathologies} & \textbf{Base LLM}              & \textbf{RadPrompt}            & \textbf{Base LLM}             & \textbf{RadPrompt}            \\ \hline
Atelectasis          & \textbf{0.838{\fontsize{8}{9}\selectfont \phantom{00}(0.792, 0.878)}} & 0.830{\fontsize{8}{9}\selectfont \phantom{00}(0.787, 0.870)}          & \textbf{0.842{\fontsize{8}{9}\selectfont \phantom{00}(0.790, 0.884)}} & 0.782{\fontsize{8}{9}\selectfont \phantom{00}(0.739, 0.822)}          \\
Cardiomegaly         & \textbf{0.809{\fontsize{8}{9}\selectfont \phantom{00}(0.771, 0.842)}} & 0.790{\fontsize{8}{9}\selectfont \phantom{00}(0.740, 0.835)}          & 0.706{\fontsize{8}{9}\selectfont \phantom{00}(0.657, 0.755)}          & \textbf{0.807{\fontsize{8}{9}\selectfont \phantom{00}(0.756, 0.853)}} \\
Consolidation        & 0.588{\fontsize{8}{9}\selectfont \phantom{00}(0.507, 0.662)}          & \textbf{0.743{\fontsize{8}{9}\selectfont \phantom{00}(0.665, 0.815)}} & 0.368{\fontsize{8}{9}\selectfont \phantom{00}(0.302, 0.430)}          & \textbf{0.625{\fontsize{8}{9}\selectfont \phantom{00}(0.544, 0.700)}} \\
Edema                & 0.737{\fontsize{8}{9}\selectfont \phantom{00}(0.695, 0.778)}          & \textbf{0.794{\fontsize{8}{9}\selectfont \phantom{00}(0.752, 0.834)}} & 0.729{\fontsize{8}{9}\selectfont \phantom{00}(0.686, 0.766)}          & \textbf{0.804{\fontsize{8}{9}\selectfont \phantom{00}(0.762, 0.845)}} \\
\hline
\end{tabular}
}
\caption{Weighted F1 Scores across positive mention detection, negation detection, and uncertainty detection for RadPrompt on MIMIC-CXR gold-standard test set. The ``Base LLM'' column refers to the first-turn prediction of the LLM, and the ``RadPrompt'' column to the second-turn prediction.  The scores correspond to the averages across 1000 bootstrap replicates and are reported alongside their confidence intervals.}
\label{tab:radprompt_f1}
\end{table*}
\renewcommand{\arraystretch}{1.3}
\begin{table*}
\centering
{\small
\begin{tabular}{@{\extracolsep{6pt}}lcccc@{}}
                     \hline
                     & \multicolumn{4}{c}{\textbf{Mention Detection F1}}                                                                 \\ \cline{2-5}
                     & \multicolumn{2}{c}{\textbf{Gemini-1.5 Pro}}                   & \multicolumn{2}{c}{\textbf{Llama-2 70B}}                      \\ \cline{2-3} \cline{4-5}
\textbf{Pathologies} & \textbf{Base LLM}              & \textbf{RadPrompt}            & \textbf{Base LLM}             & \textbf{RadPrompt}            \\ \hline
Atelectasis          & 0.785{\fontsize{8}{9}\selectfont \phantom{00}(0.748, 0.819)} & \textbf{0.926{\fontsize{8}{9}\selectfont \phantom{00}(0.901, 0.949)}} & 0.746{\fontsize{8}{9}\selectfont \phantom{00}(0.706, 0.785)}          & \textbf{0.939{\fontsize{8}{9}\selectfont \phantom{00}(0.914, 0.961)}} \\
Cardiomegaly         & 0.827{\fontsize{8}{9}\selectfont \phantom{00}(0.793, 0.861)} & \textbf{0.848{\fontsize{8}{9}\selectfont \phantom{00}(0.815, 0.882)}} & 0.767{\fontsize{8}{9}\selectfont \phantom{00}(0.731, 0.803)}          & \textbf{0.865{\fontsize{8}{9}\selectfont \phantom{00}(0.832, 0.896)}} \\
Consolidation        & 0.516{\fontsize{8}{9}\selectfont \phantom{00}(0.450, 0.579)} & \textbf{0.869{\fontsize{8}{9}\selectfont \phantom{00}(0.817, 0.912)}} & 0.374{\fontsize{8}{9}\selectfont \phantom{00}(0.318, 0.428)}          & \textbf{0.738{\fontsize{8}{9}\selectfont \phantom{00}(0.672, 0.797)}} \\
Edema                & 0.781{\fontsize{8}{9}\selectfont \phantom{00}(0.745, 0.819)} & \textbf{0.869{\fontsize{8}{9}\selectfont \phantom{00}(0.838, 0.898)}} & 0.744{\fontsize{8}{9}\selectfont \phantom{00}(0.704, 0.782)}          & \textbf{0.883{\fontsize{8}{9}\selectfont \phantom{00}(0.854, 0.911)}} \\
\hline
\end{tabular}
}
\caption{Mention detection F1 Scores for RadPrompt on MIMIC-CXR gold-standard test set. The ``Base LLM'' column refers to the first-turn prediction of the LLM, and the ``RadPrompt'' column to the second-turn prediction.  The scores correspond to the averages across 1000 bootstrap replicates and are reported alongside their confidence intervals.}
\label{tab:radprompt_f1_mention}
\end{table*}
\renewcommand{\arraystretch}{1.3}
\begin{table*}
\centering
{\small
\begin{tabular}{@{\extracolsep{6pt}}lcccc@{}}
                     \hline
                     & \multicolumn{4}{c}{\textbf{Negation Detection F1}}                                                                 \\ \cline{2-5}
                     & \multicolumn{2}{c}{\textbf{Gemini-1.5 Pro}}                   & \multicolumn{2}{c}{\textbf{Llama-2 70B}}                      \\ \cline{2-3} \cline{4-5}
\textbf{Pathologies} & \textbf{Base LLM}              & \textbf{RadPrompt}            & \textbf{Base LLM}             & \textbf{RadPrompt}            \\ \hline
Atelectasis          & 0.066{\fontsize{8}{9}\selectfont \phantom{00}(0.000, 0.143)}          & \textbf{0.340{\fontsize{8}{9}\selectfont \phantom{00}(0.000, 0.616)}} & 0.015{\fontsize{8}{9}\selectfont \phantom{00}(0.000, 0.047)}          & \textbf{0.579{\fontsize{8}{9}\selectfont \phantom{00}(0.000, 0.909)}} \\
Cardiomegaly         & 0.673{\fontsize{8}{9}\selectfont \phantom{00}(0.594, 0.739)}          & \textbf{0.742{\fontsize{8}{9}\selectfont \phantom{00}(0.667, 0.811)}} & 0.546{\fontsize{8}{9}\selectfont \phantom{00}(0.477, 0.617)}          & \textbf{0.852{\fontsize{8}{9}\selectfont \phantom{00}(0.789, 0.906)}} \\
Consolidation        & 0.286{\fontsize{8}{9}\selectfont \phantom{00}(0.188, 0.379)}          & \textbf{0.790{\fontsize{8}{9}\selectfont \phantom{00}(0.654, 0.893)}} & 0.210{\fontsize{8}{9}\selectfont \phantom{00}(0.136, 0.293)}          & \textbf{0.739{\fontsize{8}{9}\selectfont \phantom{00}(0.591, 0.857)}} \\
Edema                & 0.656{\fontsize{8}{9}\selectfont \phantom{00}(0.583, 0.721)}          & \textbf{0.801{\fontsize{8}{9}\selectfont \phantom{00}(0.737, 0.857)}} & 0.555{\fontsize{8}{9}\selectfont \phantom{00}(0.483, 0.621)}          & \textbf{0.833{\fontsize{8}{9}\selectfont \phantom{00}(0.769, 0.890)}} \\
\hline
\end{tabular}
}
\caption{Negation detection F1 Scores for RadPrompt on MIMIC-CXR gold-standard test set. The ``Base LLM'' column refers to the first-turn prediction of the LLM, and the ``RadPrompt'' column to the second-turn prediction.  The scores correspond to the averages across 1000 bootstrap replicates and are reported alongside their confidence intervals.}
\label{tab:radprompt_f1_negation}
\end{table*}
\renewcommand{\arraystretch}{1.3}
\begin{table*}
\centering
{\small
\begin{tabular}{@{\extracolsep{6pt}}lcccc@{}}
                     \hline
                     & \multicolumn{4}{c}{\textbf{Uncertainty Detection F1}}                                                                 \\ \cline{2-5}
                     & \multicolumn{2}{c}{\textbf{Gemini-1.5 Pro}}                   & \multicolumn{2}{c}{\textbf{Llama-2 70B}}                      \\ \cline{2-3} \cline{4-5}
\textbf{Pathologies} & \textbf{Base LLM}              & \textbf{RadPrompt}            & \textbf{Base LLM}             & \textbf{RadPrompt}            \\ \hline
Atelectasis          & 0.301{\fontsize{8}{9}\selectfont \phantom{00}(0.136, 0.464)}          & \textbf{0.376{\fontsize{8}{9}\selectfont \phantom{00}(0.208, 0.536)}} & 0.364{\fontsize{8}{9}\selectfont \phantom{00}(0.143, 0.560)}          & \textbf{0.386{\fontsize{8}{9}\selectfont \phantom{00}(0.256, 0.515)}} \\
Cardiomegaly         & \textbf{0.385{\fontsize{8}{9}\selectfont \phantom{00}(0.235, 0.529)}} & 0.170{\fontsize{8}{9}\selectfont \phantom{00}(0.044, 0.320)}          & 0.000{\fontsize{8}{9}\selectfont \phantom{00}(0.000, 0.000)}          & \textbf{0.095{\fontsize{8}{9}\selectfont \phantom{00}(0.000, 0.227)}} \\
Consolidation        & 0.258{\fontsize{8}{9}\selectfont \phantom{00}(0.138, 0.386)}          & \textbf{0.448{\fontsize{8}{9}\selectfont \phantom{00}(0.250, 0.643)}} & 0.236{\fontsize{8}{9}\selectfont \phantom{00}(0.133, 0.341)}          & \textbf{0.542{\fontsize{8}{9}\selectfont \phantom{00}(0.367, 0.706)}} \\
Edema                & 0.253{\fontsize{8}{9}\selectfont \phantom{00}(0.082, 0.410)}          & \textbf{0.317{\fontsize{8}{9}\selectfont \phantom{00}(0.087, 0.522)}} & 0.382{\fontsize{8}{9}\selectfont \phantom{00}(0.154, 0.571)}          & \textbf{0.382{\fontsize{8}{9}\selectfont \phantom{00}(0.148, 0.585)}} \\
\hline
\end{tabular}
}
\caption{Uncertainty detection F1 Scores for RadPrompt on MIMIC-CXR gold-standard test set. The ``Base LLM'' column refers to the first-turn prediction of the LLM, and the ``RadPrompt'' column to the second-turn prediction.  The scores correspond to the averages across 1000 bootstrap replicates and are reported alongside their confidence intervals.}
\label{tab:radprompt_f1_uncertainty}
\end{table*}
\renewcommand{\arraystretch}{1.3}
\begin{table*}
\centering
{\small
\begin{tabular}{@{\extracolsep{6pt}}lcccc@{}}
                     \hline
                     & \multicolumn{4}{c}{\textbf{Positive Mention Detection F1}}                                                                 \\ \cline{2-5}
                     & \multicolumn{2}{c}{\textbf{Gemini-1.5 Pro}}                   & \multicolumn{2}{c}{\textbf{Llama-2 70B}}                      \\ \cline{2-3} \cline{4-5}
\textbf{Pathologies} & \textbf{Base LLM}              & \textbf{RadPrompt}            & \textbf{Base LLM}             & \textbf{RadPrompt}            \\ \hline
Atelectasis          & \textbf{0.900{\fontsize{8}{9}\selectfont \phantom{00}(0.869, 0.928)}} & 0.878{\fontsize{8}{9}\selectfont \phantom{00}(0.843, 0.910)}          & \textbf{0.899{\fontsize{8}{9}\selectfont \phantom{00}(0.866, 0.927)}} & 0.819{\fontsize{8}{9}\selectfont \phantom{00}(0.775, 0.859)}          \\
Cardiomegaly         & \textbf{0.928{\fontsize{8}{9}\selectfont \phantom{00}(0.895, 0.957)}} & 0.879{\fontsize{8}{9}\selectfont \phantom{00}(0.840, 0.914)}          & 0.869{\fontsize{8}{9}\selectfont \phantom{00}(0.828, 0.905)}          & \textbf{0.850{\fontsize{8}{9}\selectfont \phantom{00}(0.805, 0.892)}} \\
Consolidation        & \textbf{0.811{\fontsize{8}{9}\selectfont \phantom{00}(0.729, 0.881)}} & 0.810{\fontsize{8}{9}\selectfont \phantom{00}(0.733, 0.883)}          & 0.469{\fontsize{8}{9}\selectfont \phantom{00}(0.382, 0.550)}          & \textbf{0.597{\fontsize{8}{9}\selectfont \phantom{00}(0.496, 0.689)}} \\
Edema                & \textbf{0.824{\fontsize{8}{9}\selectfont \phantom{00}(0.775, 0.867)}} & 0.822{\fontsize{8}{9}\selectfont \phantom{00}(0.773, 0.867)}          & \textbf{0.866{\fontsize{8}{9}\selectfont \phantom{00}(0.822, 0.906)}} & 0.816{\fontsize{8}{9}\selectfont \phantom{00}(0.767, 0.866)}          \\
\hline
\end{tabular}
}
\caption{Positive mention detection F1 Scores for RadPrompt on MIMIC-CXR gold-standard test set. The ``Base LLM'' column refers to the first-turn prediction of the LLM, and the ``RadPrompt'' column to the second-turn prediction. The scores correspond to the averages across 1000 bootstrap replicates and are reported alongside their confidence intervals.}
\label{tab:radprompt_f1_positive}
\end{table*}

\renewcommand{\arraystretch}{1.3}
\begin{table*}
\centering
{\small
%
} \\ \hline
Atelectasis          & 0.830{\fontsize{8}{9}\selectfont \phantom{00}(0.767, 0.888)}                                                             &\phantom{-0} 37.9 \phantom{-0-0}(22.9, 56.8)                                                                       &\phantom{00} -7.1 \phantom{000}(-11.6, -3.0)                                                               \\
Cardiomegaly         & 0.810{\fontsize{8}{9}\selectfont \phantom{00}(0.747, 0.867)}                                                             &\phantom{-0} 41.3 \phantom{-0-0}(28.6, 57.9)                                                                       &\phantom{0} -11.0 \phantom{000}(-16.6, -6.0)                                                              \\
Consolidation        & 0.929{\fontsize{8}{9}\selectfont \phantom{00}(0.903, 0.953)}                                                             &\phantom{-0} 27.7 \phantom{-0-0}(21.7, 34.8)                                                                       &\phantom{00} -2.3 \phantom{0000}(-4.2, -0.7)                                                                \\
Edema                & 0.529{\fontsize{8}{9}\selectfont \phantom{00}(0.381, 0.639)}                                                             &\phantom{-0} 41.5 \phantom{00-0}(-4.1, 99.8)                                                                       &\phantom{0} -15.1 \phantom{000}(-27.3, -0.8)                                                              \\
Enlarged Card.       & 0.844{\fontsize{8}{9}\selectfont \phantom{00}(0.790, 0.894)}                                                             &\phantom{-0} Inf. \phantom{-00--0}(Inf., Inf.)                                                                       &\phantom{00} -7.0 \phantom{000}(-10.6, -3.3)                                                               \\
Fracture             & 0.684{\fontsize{8}{9}\selectfont \phantom{00}(0.531, 0.817)}                                                             &\phantom{-0} 12.6 \phantom{00-0}(-5.8, 38.8)                                                                       &\phantom{0} -10.3 \phantom{000}(-20.0, -0.6)                                                              \\
Lung Lesion          & 0.699{\fontsize{8}{9}\selectfont \phantom{00}(0.577, 0.817)}                                                             &\phantom{-} 191.8 \phantom{--}(132.3, 268.1)                                                                    &\phantom{0} -14.3 \phantom{000}(-23.3, -5.7)                                                              \\
Lung Opacity         & 0.692{\fontsize{8}{9}\selectfont \phantom{00}(0.636, 0.748)}                                                             &\phantom{-00} 2.9 \phantom{00-0}(-6.5, 13.4)                                                                        &\phantom{00} -2.8 \phantom{0000}(-5.7, -0.1)                                                                \\
Pleur. Effusion      & 0.615{\fontsize{8}{9}\selectfont \phantom{00}(0.562, 0.665)}                                                             &\phantom{0} -22.6 \phantom{00}(-29.5, -16.1)                                                                    &\phantom{00} -3.9 \phantom{0000}(-7.6, -0.9)                                                                \\
Pleur. Other         & 0.106{\fontsize{8}{9}\selectfont \phantom{00}(0.059, 0.163)}                                                             &\phantom{0} -80.9 \phantom{00}(-89.5, -70.8)                                                                    &\phantom{-0} 34.2 \phantom{00-}(-8.5, 104.7)                                                               \\
Pneumonia            & 0.519{\fontsize{8}{9}\selectfont \phantom{00}(0.374, 0.654)}                                                             &\phantom{-} 259.1 \phantom{--}(144.6, 433.0)                                                                    &\phantom{0} -21.0 \phantom{00}(-33.4, -10.7)                                                             \\
Pneumothorax         & 0.606{\fontsize{8}{9}\selectfont \phantom{00}(0.550, 0.661)}                                                             &\phantom{0} -16.0 \phantom{000}(-23.4, -8.2)                                                                     &\phantom{00} -3.3 \phantom{0000}(-6.0, -0.2)                                                                \\
Sup. Devices         & 0.822{\fontsize{8}{9}\selectfont \phantom{00}(0.785, 0.857)}                                                             &\phantom{-00} 6.2 \phantom{00-0}(-0.3, 13.5)                                                                        &\phantom{00} -4.2 \phantom{0000}(-6.3, -2.3)                                                                \\ \hline
Macro Avg.           & 0.668{\fontsize{8}{9}\selectfont \phantom{00}(0.639, 0.694)}                                                             &\phantom{-0} 27.4 \phantom{-0-0}(21.9, 32.6)                                                                       &\phantom{00} -8.0 \phantom{000}(-10.2, -5.7)                                                               \\
Weighted Avg.        & 0.695{\fontsize{8}{9}\selectfont \phantom{00}(0.668, 0.748)}                                                             &\phantom{-00} 3.0 \phantom{00-0}(-1.3, 10.4)                                                                        &\phantom{0} -11.7 \phantom{000}(-13.3, -5.3)                                                              \\ \hline

\end{tabular}
}
\caption{Weighted average F1 scores for Llama-2-based RadPrompt on the CUH test set, alongside improvements over  1\textsuperscript{st} turn Llama-2 and RadPert predictions.  The scores correspond to the averages across 1000 bootstrap replicates and are reported alongside their confidence intervals.}
\label{tab:radprompt_llama2_cuh}
\end{table*}
\renewcommand{\arraystretch}{1.3}
\begin{table*}
\centering
{\small
\begin{tabular}{@{\extracolsep{6pt}}lcccc@{}}
\hline
                     & \multicolumn{4}{c}{\textbf{Sub-Task F1 Scores}}                                                                               \\ \cline{2-5} 
                     & \multicolumn{2}{c}{\textbf{Negation Detection}}               & \multicolumn{2}{c}{\textbf{Uncertainty Detection}}            \\ \cline{2-3} \cline{4-5}
\textbf{Pathologies} & \textbf{Base Llama-2}         & \textbf{RadPrompt}            & \textbf{Base Llama-2}         & \textbf{RadPrompt}            \\ \hline
Atelectasis          & 0.175{\fontsize{8}{9}\selectfont \phantom{00}(0.111, 0.238)}          & \textbf{0.853{\fontsize{8}{9}\selectfont \phantom{00}(0.766, 0.923)}} & 0.000{\fontsize{8}{9}\selectfont \phantom{00}(0.000, 0.000)}          & \textbf{0.126{\fontsize{8}{9}\selectfont \phantom{00}(0.000, 0.400)}} \\
Cardiomegaly         & 0.579{\fontsize{8}{9}\selectfont \phantom{00}(0.515, 0.639)}          & \textbf{0.835{\fontsize{8}{9}\selectfont \phantom{00}(0.779, 0.884)}} & 0.000{\fontsize{8}{9}\selectfont \phantom{00}(0.000, 0.000)}          & \textbf{0.412{\fontsize{8}{9}\selectfont \phantom{00}(0.000, 0.727)}} \\
Consolidation        & 0.665{\fontsize{8}{9}\selectfont \phantom{00}(0.608, 0.720)}          & \textbf{0.923{\fontsize{8}{9}\selectfont \phantom{00}(0.887, 0.953)}} & 0.145{\fontsize{8}{9}\selectfont \phantom{00}(0.000, 0.298)}          & \textbf{0.490{\fontsize{8}{9}\selectfont \phantom{00}(0.154, 0.769)}} \\
Edema                & 0.160{\fontsize{8}{9}\selectfont \phantom{00}(0.102, 0.223)}          & \textbf{0.444{\fontsize{8}{9}\selectfont \phantom{00}(0.278, 0.597)}} & 0.322{\fontsize{8}{9}\selectfont \phantom{00}(0.000, 1.000)}          & \textbf{0.408{\fontsize{8}{9}\selectfont \phantom{00}(0.000, 0.800)}} \\
Enlarged Card.       & 0.000{\fontsize{8}{9}\selectfont \phantom{00}(0.000, 0.000)}          & \textbf{0.904{\fontsize{8}{9}\selectfont \phantom{00}(0.854, 0.947)}} & 0.000{\fontsize{8}{9}\selectfont \phantom{00}(0.000, 0.000)}          & \textbf{0.639{\fontsize{8}{9}\selectfont \phantom{00}(0.471, 0.791)}} \\
Fracture             & 0.052{\fontsize{8}{9}\selectfont \phantom{00}(0.018, 0.098)}          & \textbf{0.269{\fontsize{8}{9}\selectfont \phantom{00}(0.071, 0.483)}} & 0.000{\fontsize{8}{9}\selectfont \phantom{00}(0.000, 0.000)}          & 0.000{\fontsize{8}{9}\selectfont \phantom{00}(0.000, 0.000)}          \\
Lung Lesion          & 0.285{\fontsize{8}{9}\selectfont \phantom{00}(0.215, 0.359)}          & \textbf{0.790{\fontsize{8}{9}\selectfont \phantom{00}(0.684, 0.884)}} & 0.000{\fontsize{8}{9}\selectfont \phantom{00}(0.000, 0.000)}          & 0.000{\fontsize{8}{9}\selectfont \phantom{00}(0.000, 0.000)}          \\
Lung Opacity         & 0.022{\fontsize{8}{9}\selectfont \phantom{00}(0.000, 0.056)}          & \textbf{0.187{\fontsize{8}{9}\selectfont \phantom{00}(0.000, 0.421)}} & \textbf{0.228{\fontsize{8}{9}\selectfont \phantom{00}(0.000, 0.667)}} & 0.000{\fontsize{8}{9}\selectfont \phantom{00}(0.000, 0.000)}          \\
Pleural Effusion     & \textbf{0.758{\fontsize{8}{9}\selectfont \phantom{00}(0.717, 0.797)}} & 0.532{\fontsize{8}{9}\selectfont \phantom{00}(0.468, 0.592)}          & \textbf{0.414{\fontsize{8}{9}\selectfont \phantom{00}(0.000, 0.800)}} & 0.319{\fontsize{8}{9}\selectfont \phantom{00}(0.000, 0.600)}          \\
Pleural Other        & \textbf{0.556{\fontsize{8}{9}\selectfont \phantom{00}(0.490, 0.615)}} & 0.035{\fontsize{8}{9}\selectfont \phantom{00}(0.000, 0.077)}          & 0.000{\fontsize{8}{9}\selectfont \phantom{00}(0.000, 0.000)}          & \textbf{0.515{\fontsize{8}{9}\selectfont \phantom{00}(0.000, 1.000)}} \\
Pneumonia            & 0.113{\fontsize{8}{9}\selectfont \phantom{00}(0.063, 0.171)}          & \textbf{0.642{\fontsize{8}{9}\selectfont \phantom{00}(0.424, 0.813)}} & 0.128{\fontsize{8}{9}\selectfont \phantom{00}(0.028, 0.237)}          & \textbf{0.310{\fontsize{8}{9}\selectfont \phantom{00}(0.087, 0.522)}} \\
Pneumothorax         & \textbf{0.730{\fontsize{8}{9}\selectfont \phantom{00}(0.683, 0.771)}} & 0.610{\fontsize{8}{9}\selectfont \phantom{00}(0.551, 0.663)}          & 0.000{\fontsize{8}{9}\selectfont \phantom{00}(0.000, 0.000)}          & 0.000{\fontsize{8}{9}\selectfont \phantom{00}(0.000, 0.000)}          \\
Support Devices      & 0.000{\fontsize{8}{9}\selectfont \phantom{00}(0.000, 0.000)}          & 0.000{\fontsize{8}{9}\selectfont \phantom{00}(0.000, 0.000)}          & 0.000{\fontsize{8}{9}\selectfont \phantom{00}(0.000, 0.000)}          & 0.000{\fontsize{8}{9}\selectfont \phantom{00}(0.000, 0.000)}          \\ \hline
Macro Average        & 0.377{\fontsize{8}{9}\selectfont \phantom{00}(0.356, 0.413)}          & \textbf{0.596{\fontsize{8}{9}\selectfont \phantom{00}(0.550, 0.664)}} & 0.296{\fontsize{8}{9}\selectfont \phantom{00}(0.149, 0.441)}          & \textbf{0.459{\fontsize{8}{9}\selectfont \phantom{00}(0.335, 0.585)}} \\
Weighted Average     & 0.617{\fontsize{8}{9}\selectfont \phantom{00}(0.588, 0.643)}          & \textbf{0.607{\fontsize{8}{9}\selectfont \phantom{00}(0.568, 0.705)}} & 0.263{\fontsize{8}{9}\selectfont \phantom{00}(0.127, 0.413)}          & \textbf{0.506{\fontsize{8}{9}\selectfont \phantom{00}(0.387, 0.616)}} \\ \hline
                     & \multicolumn{2}{c}{\textbf{Positive Mention Detection}}       & \multicolumn{2}{c}{\textbf{Mention Detection}}                \\ \cline{2-3} \cline{4-5}
\textbf{Pathologies} & \textbf{Base Llama-2}         & \textbf{RadPrompt}            & \textbf{Base Llama-2}         & \textbf{RadPrompt}            \\ \hline
Atelectasis          & \textbf{0.889{\fontsize{8}{9}\selectfont \phantom{00}(0.826, 0.938)}} & 0.843{\fontsize{8}{9}\selectfont \phantom{00}(0.779, 0.902)}          & 0.454{\fontsize{8}{9}\selectfont \phantom{00}(0.394, 0.510)}          & \textbf{0.868{\fontsize{8}{9}\selectfont \phantom{00}(0.822, 0.908)}} \\
Cardiomegaly         & 0.885{\fontsize{8}{9}\selectfont \phantom{00}(0.750, 0.976)}          & \textbf{0.888{\fontsize{8}{9}\selectfont \phantom{00}(0.765, 0.977)}} & 0.625{\fontsize{8}{9}\selectfont \phantom{00}(0.567, 0.679)}          & \textbf{0.851{\fontsize{8}{9}\selectfont \phantom{00}(0.805, 0.895)}} \\
Consolidation        & 0.806{\fontsize{8}{9}\selectfont \phantom{00}(0.761, 0.851)}          & \textbf{0.950{\fontsize{8}{9}\selectfont \phantom{00}(0.924, 0.975)}} & 0.737{\fontsize{8}{9}\selectfont \phantom{00}(0.704, 0.771)}          & \textbf{0.966{\fontsize{8}{9}\selectfont \phantom{00}(0.951, 0.980)}} \\
Edema                & \textbf{0.828{\fontsize{8}{9}\selectfont \phantom{00}(0.631, 0.960)}} & 0.697{\fontsize{8}{9}\selectfont \phantom{00}(0.400, 0.917)}          & 0.230{\fontsize{8}{9}\selectfont \phantom{00}(0.167, 0.295)}          & \textbf{0.546{\fontsize{8}{9}\selectfont \phantom{00}(0.405, 0.659)}} \\
Enlarged Card.       & 0.000{\fontsize{8}{9}\selectfont \phantom{00}(0.000, 0.000)}          & 0.000{\fontsize{8}{9}\selectfont \phantom{00}(0.000, 0.000)}          & 0.026{\fontsize{8}{9}\selectfont \phantom{00}(0.000, 0.065)}          & \textbf{0.876{\fontsize{8}{9}\selectfont \phantom{00}(0.829, 0.919)}} \\
Fracture             & 0.843{\fontsize{8}{9}\selectfont \phantom{00}(0.711, 0.947)}          & \textbf{0.847{\fontsize{8}{9}\selectfont \phantom{00}(0.722, 0.955)}} & 0.196{\fontsize{8}{9}\selectfont \phantom{00}(0.136, 0.257)}          & \textbf{0.637{\fontsize{8}{9}\selectfont \phantom{00}(0.500, 0.761)}} \\
Lung Lesion          & 0.094{\fontsize{8}{9}\selectfont \phantom{00}(0.021, 0.172)}          & \textbf{0.434{\fontsize{8}{9}\selectfont \phantom{00}(0.000, 0.750)}} & 0.204{\fontsize{8}{9}\selectfont \phantom{00}(0.154, 0.257)}          & \textbf{0.742{\fontsize{8}{9}\selectfont \phantom{00}(0.635, 0.837)}} \\
Lung Opacity         & 0.701{\fontsize{8}{9}\selectfont \phantom{00}(0.655, 0.746)}          & \textbf{0.716{\fontsize{8}{9}\selectfont \phantom{00}(0.659, 0.772)}} & 0.523{\fontsize{8}{9}\selectfont \phantom{00}(0.476, 0.566)}          & \textbf{0.696{\fontsize{8}{9}\selectfont \phantom{00}(0.638, 0.755)}} \\
Pleural Effusion     & \textbf{0.916{\fontsize{8}{9}\selectfont \phantom{00}(0.875, 0.953)}} & 0.848{\fontsize{8}{9}\selectfont \phantom{00}(0.789, 0.901)}          & \textbf{0.831{\fontsize{8}{9}\selectfont \phantom{00}(0.801, 0.859)}} & 0.711{\fontsize{8}{9}\selectfont \phantom{00}(0.665, 0.752)}          \\
Pleural Other        & 0.687{\fontsize{8}{9}\selectfont \phantom{00}(0.451, 0.875)}          & \textbf{0.836{\fontsize{8}{9}\selectfont \phantom{00}(0.640, 0.968)}} & \textbf{0.577{\fontsize{8}{9}\selectfont \phantom{00}(0.517, 0.635)}} & 0.162{\fontsize{8}{9}\selectfont \phantom{00}(0.096, 0.239)}          \\
Pneumonia            & 0.187{\fontsize{8}{9}\selectfont \phantom{00}(0.077, 0.298)}          & \textbf{0.479{\fontsize{8}{9}\selectfont \phantom{00}(0.240, 0.684)}} & 0.147{\fontsize{8}{9}\selectfont \phantom{00}(0.099, 0.193)}          & \textbf{0.580{\fontsize{8}{9}\selectfont \phantom{00}(0.454, 0.693)}} \\
Pneumothorax         & \textbf{0.618{\fontsize{8}{9}\selectfont \phantom{00}(0.333, 0.833)}} & 0.607{\fontsize{8}{9}\selectfont \phantom{00}(0.286, 0.857)}          & \textbf{0.734{\fontsize{8}{9}\selectfont \phantom{00}(0.691, 0.777)}} & 0.625{\fontsize{8}{9}\selectfont \phantom{00}(0.568, 0.678)}          \\
Support Devices      & 0.780{\fontsize{8}{9}\selectfont \phantom{00}(0.736, 0.819)}          & \textbf{0.828{\fontsize{8}{9}\selectfont \phantom{00}(0.792, 0.863)}} & 0.646{\fontsize{8}{9}\selectfont \phantom{00}(0.602, 0.688)}          & \textbf{0.818{\fontsize{8}{9}\selectfont \phantom{00}(0.782, 0.852)}} \\ \hline
Macro Average        & 0.687{\fontsize{8}{9}\selectfont \phantom{00}(0.647, 0.723)}          & \textbf{0.752{\fontsize{8}{9}\selectfont \phantom{00}(0.696, 0.798)}} & 0.461{\fontsize{8}{9}\selectfont \phantom{00}(0.443, 0.499)}          & \textbf{0.698{\fontsize{8}{9}\selectfont \phantom{00}(0.671, 0.723)}} \\
Weighted Average     & 0.781{\fontsize{8}{9}\selectfont \phantom{00}(0.763, 0.801)}          & \textbf{0.822{\fontsize{8}{9}\selectfont \phantom{00}(0.799, 0.842)}} & 0.612{\fontsize{8}{9}\selectfont \phantom{00}(0.590, 0.652)}          & \textbf{0.726{\fontsize{8}{9}\selectfont \phantom{00}(0.704, 0.748)}} \\ \hline
\end{tabular}
}
\caption{F1 Scores for all sub-tasks for Llama-2-based RadPrompt on the CUH dataset. The ``Base Llama-2'' column refers to the first-turn prediction of the LLM, and the ``RadPrompt'' column to the second-turn prediction.  The scores correspond to the averages across 1000 bootstrap replicates and are reported alongside their confidence intervals.}
\label{tab:llama_cuh_fine}
\end{table*}


\renewcommand{\arraystretch}{1.3}
\begin{table*}
\centering
\begin{tabularx}{\textwidth}{|X|X|}
\hline
\textbf{First Turn Prompt} & \textbf{Second Turn Prompt} \\
\hline
Please accurately classify radiology reports for the presence or absence of findings. For each report, you will classify for the presence or absence of  the following findings: Enlarged Cardiomediastinum, Cardiomegaly, ....

\,

Structure your answer like the template I provided to you delimited by  triple backticks and return this template and nothing else.  

\,

ALWAYS RETURN THE FULL TEMPLATE:

\texttt{\textasciigrave \textasciigrave \textasciigrave}  \{``Enlarged Cardiomediastinum'':

\quad\quad\quad\quad {[}ANSWER{]},

\phantom{``` \{}``Cardiomegaly'':

\quad\quad\quad\quad {[}ANSWER{]}, ...

\} \texttt{\textasciigrave \textasciigrave \textasciigrave} 

\,

If the existence of a finding is mentioned, answer ``Yes''.

If a finding is mentioned as not existing, answer ``No''. 

If it cannot be determined if the patient has the findings, answer ``Maybe''.

If a finding is not mentioned in the report, answer  `Undefined''. 

\,

Important steps to consider: 

1. Read the radiology report and identify any mentions of Enlarged Cardiomediastinum, Cardiomegaly, ... 

2. For every mention, determine if it is a positive, a negative, or an uncertain one.

3. If a finding is not mentioned in the report, answer ``Undefined''.

4. For every finding, answer ``Yes'' if it is mentioned as existing (positive), ``Maybe'' if it is mentioned as uncertain, and ``No'' if it is mentioned as not existing (negative).

\,

Classify the following radiology report according to the template.  Always output the full template, even if a finding is not mentioned.

\,

\textless{}START OF REPORT\textgreater

... 

\textless{}END OF REPORT\textgreater  

\textless{}ANSWER:\textgreater{}  & 

I am using a rule-based expert model to verify your answer. Here are some insights. However, those suggestions may be wrong. Please give me your new answer after either accepting or rejecting some or all of these suggestions:

\, 

1.  The tool agrees that the overall report should be classified as ``Yes'' for Pneumonia.

2. In agreement with your previous answer, the tool detected no mentions of Enlarged Cardiomediastinum, Cardiomegaly,...

3. The tool did not detect any explicit mentions for Lung Lesion and, thus, its suggested output is ``Undefined'' for Lung Lesion.

4. The tool considers Atelectasis as ``Maybe'' because of the sentence ``...''. However, you previously classified the overall report as ``Yes'' for Atelectasis.

\,

Please use the same template for your revised answer:

\texttt{\textasciigrave \textasciigrave \textasciigrave} \{``Enlarged Cardiomediastinum'':

\quad\quad\quad\quad {[}ANSWER{]},

\phantom{``` \{}``Cardiomegaly'':

\quad\quad\quad\quad {[}ANSWER{]}, ...

\} \texttt{\textasciigrave \textasciigrave \textasciigrave} 

\\
\hline
\end{tabularx}
\caption{Example of RadPrompt first and second-turn prompts. The first-turn prompts are adapted from \citep{dorfner2024open}.}
\label{tab:prompt}
\end{table*}

\renewcommand{\arraystretch}{1.3}
\begin{table*}
{\small
\begin{tabular}{@{\extracolsep{6pt}}lcccccccc@{}}
\hline
                     & \multicolumn{4}{c}{\textbf{MIMIC-CXR Gold-Standard}}                       & \multicolumn{4}{c}{\textbf{CUH}}                               \\ \cline{2-5} \cline{6-9} 
\textbf{Pathologies} & \textbf{Null} & \textbf{Negative} & \textbf{Uncertain} & \textbf{Positive} & \textbf{Null}            & \textbf{Negative} & \textbf{Uncertain} & \textbf{Positive} \\ \hline
Atelectasis          & 469           & 4                 & 17                 & 197               & 538                      & 41                & 3                  & 68                \\
Cardiomegaly         & 452           & 82                & 14                 & 139               & 523                      & 100               & 10                 & 17                \\
Consolidation        & 592           & 23                & 17                 & 55                & \multicolumn{1}{c}{355} & 138               & 6                  & 151               \\
Edema                & 460           & 85                & 10                 & 132               & 614                      & 23                & 2                  & 11                \\
Enlarged Card.       & 617           & 28                & 1                  & 41                & 536                      & 90                & 23                 & 1                 \\
Fracture             & 637           & 8                 & 2                  & 40                & 623                      & 8                 & 0                  & 19                \\
Lung Lesion          & 621           & 4                 & 8                  & 54                & 607                      & 34                & 2                  & 7                 \\
Lung Opacity         & 493           & 23                & 0                  & 171               & 471                      & 7                 & 1                  & 171               \\
Pleural Effusion     & 317           & 82                & 18                 & 270               & 311                      & 240               & 6                  & 93                \\
Pleural Other        & 660           & 2                 & 0                  & 25                & 476                      & 158               & 2                  & 14                \\
Pneumonia            & 464           & 83                & 62                 & 78                & 617                      & 14                & 8                  & 11                \\
Pneumothorax         & 461           & 179               & 8                  & 39                & 403                      & 237               & 2                  & 8                 \\
Support Devices      & 453           & 5                 & 0                  & 229               & 369                      & 1                 & 1                  & 279               \\ \hline
Total                & 6696          & 608               & 157                & 1470              & 6443                     & 1091              & 66                 & 850               \\ \hline
\end{tabular}
}
\caption{Number of output classes per pathology for the MIMIC-CXR gold-standard test set and CUH dataset.}
\label{tab:datasets}
\end{table*}

\begin{figure*}[!ht]
    \centering
    \begin{subfigure}{.47\textwidth}
        \centering
        \includegraphics[scale=0.26]{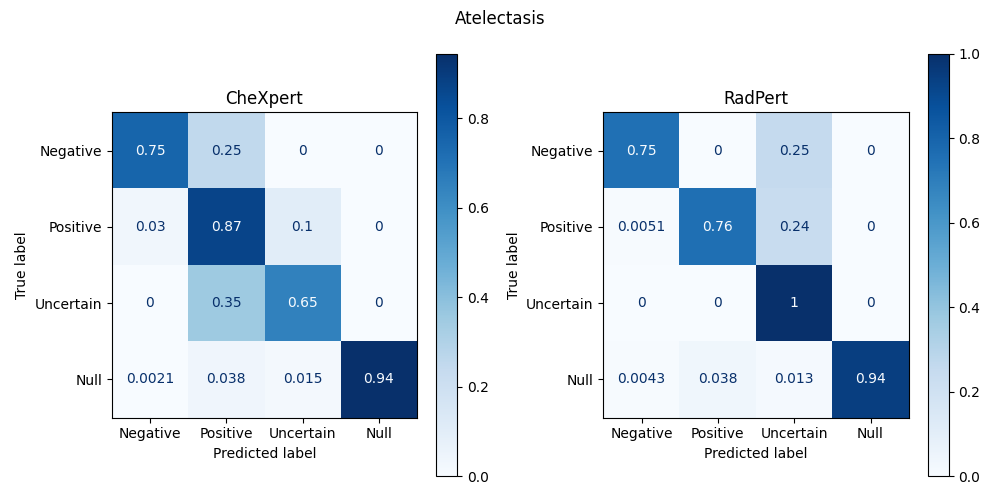}
    \end{subfigure}
    \hfill
    \begin{subfigure}{.47\textwidth}
        \centering
        \includegraphics[scale=0.26]{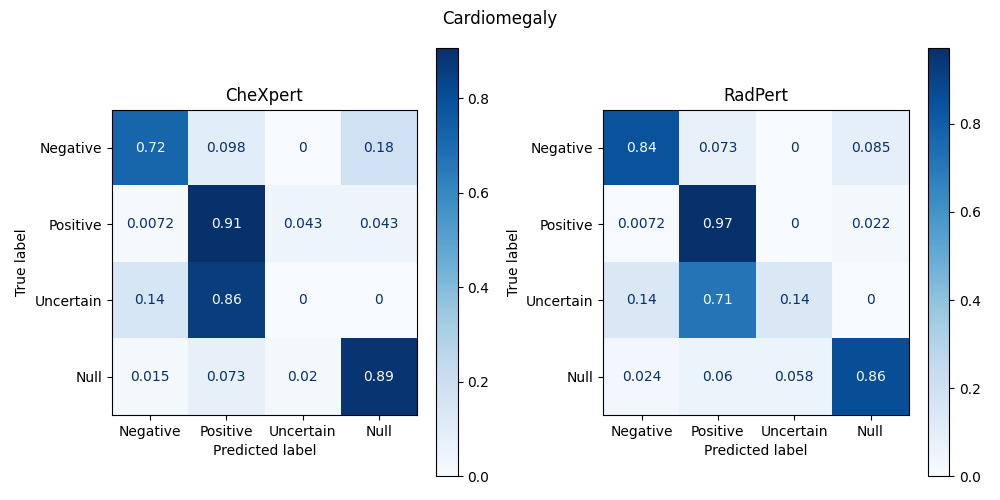}
    \end{subfigure}
    \begin{subfigure}{.47\textwidth}
        \centering
        \includegraphics[scale=0.26]{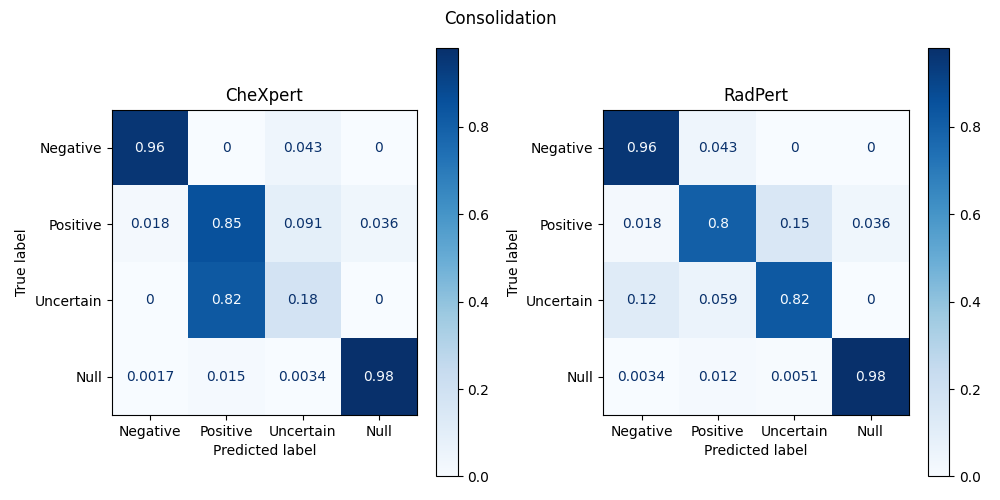}
    \end{subfigure}
    \hfill
    \begin{subfigure}{.47\textwidth}
        \centering
        \includegraphics[scale=0.26]{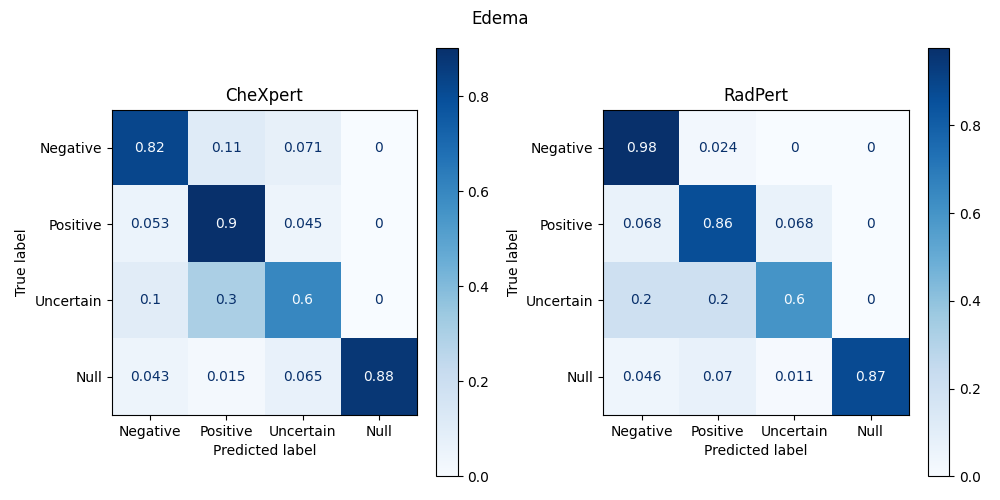}
    \end{subfigure}
    \begin{subfigure}{.47\textwidth}
        \centering
        \includegraphics[scale=0.26]{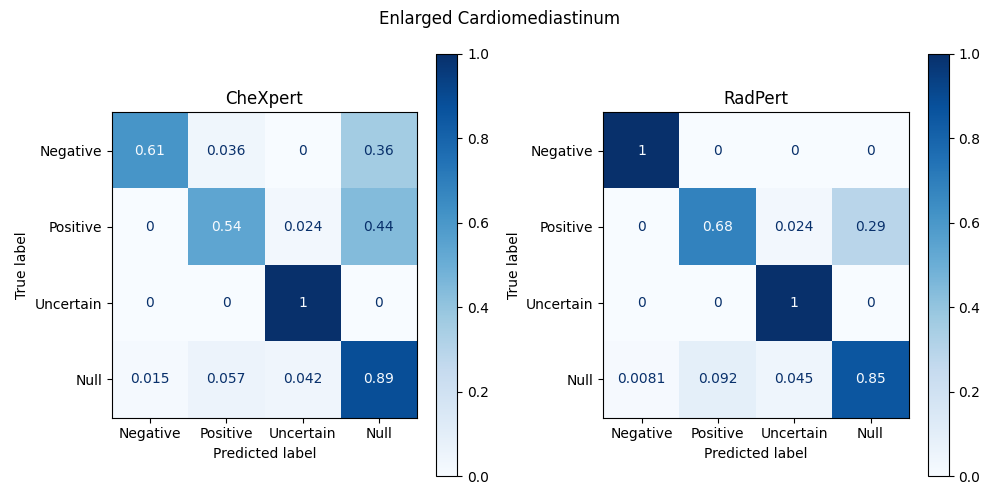}
    \end{subfigure}
    \hfill
    \begin{subfigure}{.47\textwidth}
        \centering
        \includegraphics[scale=0.26]{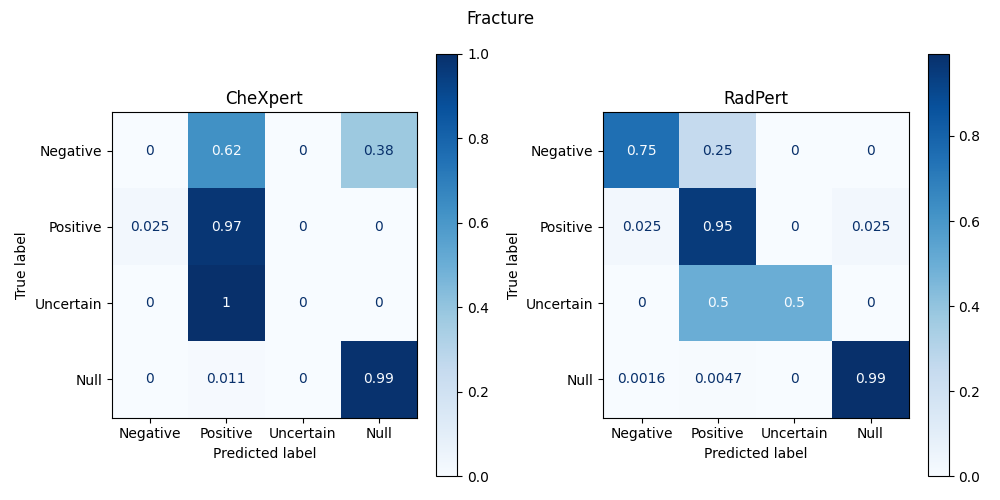}
    \end{subfigure}
    \begin{subfigure}{.47\textwidth}
        \centering
        \includegraphics[scale=0.26]{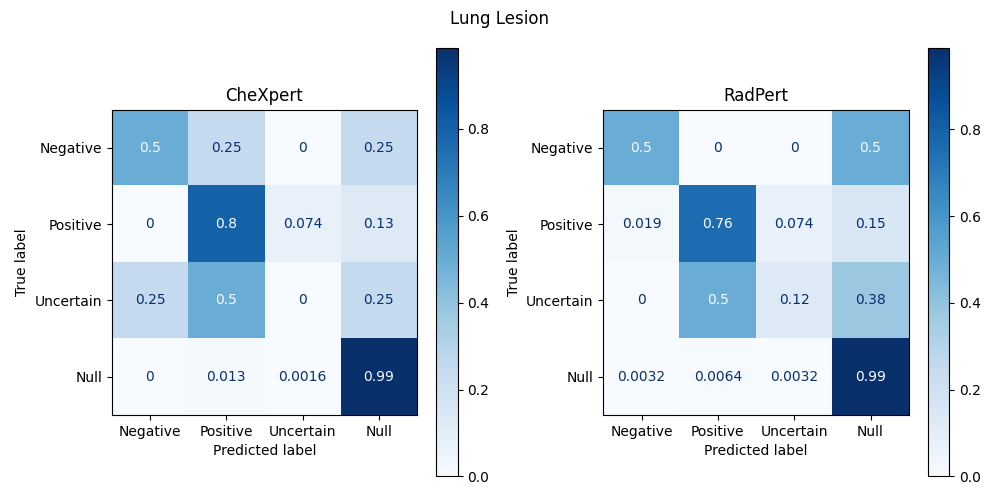}
    \end{subfigure}
    \hfill
    \begin{subfigure}{.47\textwidth}
        \centering
        \includegraphics[scale=0.26]{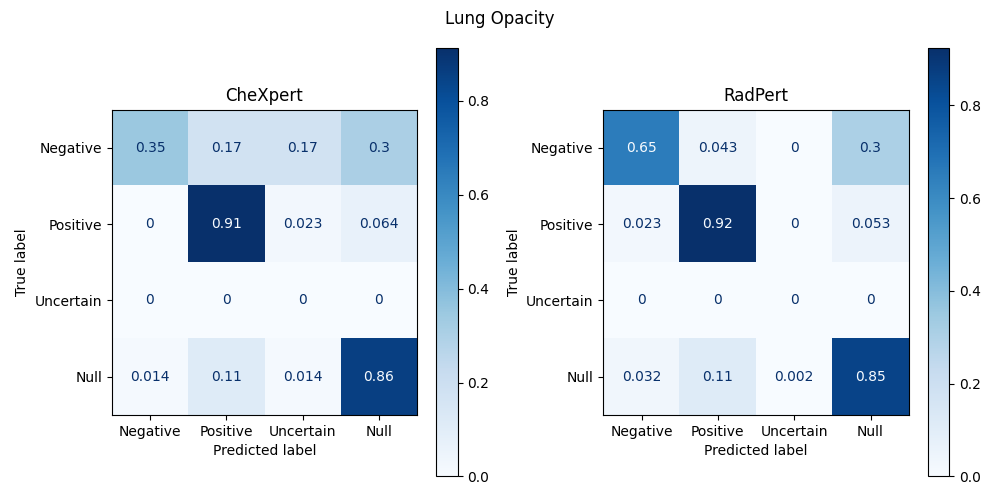}
    \end{subfigure}
    \begin{subfigure}{.47\textwidth}
        \centering
        \includegraphics[scale=0.26]{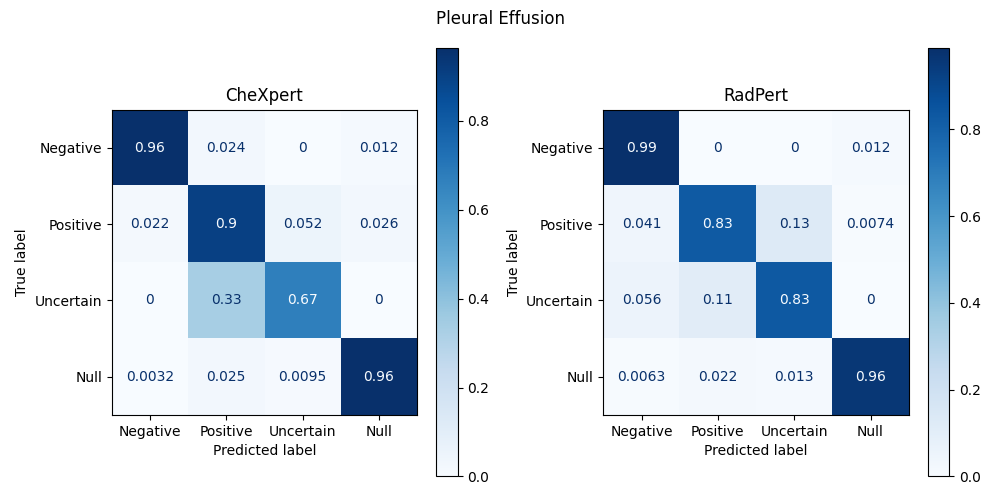}
    \end{subfigure}
    \hfill
    \begin{subfigure}{.47\textwidth}
        \centering
        \includegraphics[scale=0.26]{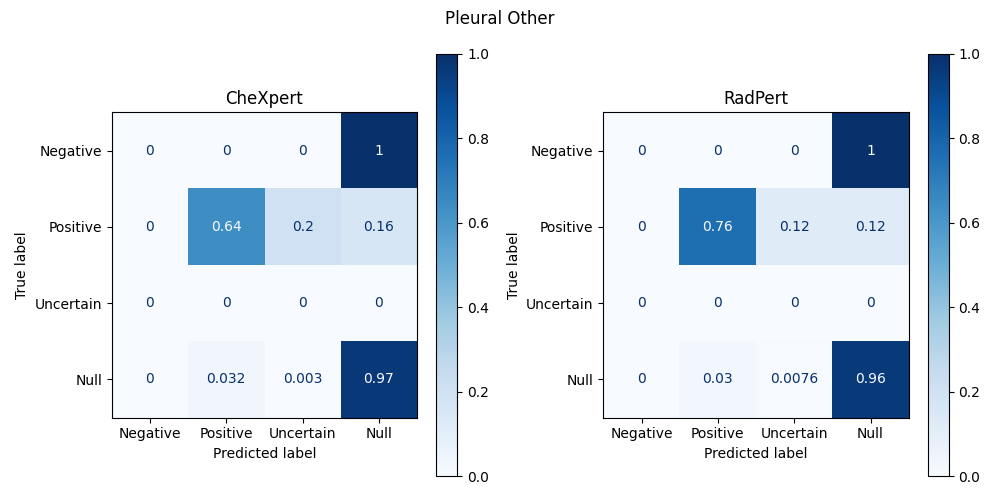}
    \end{subfigure}
    \begin{subfigure}{.47\textwidth}
        \centering
        \includegraphics[scale=0.26]{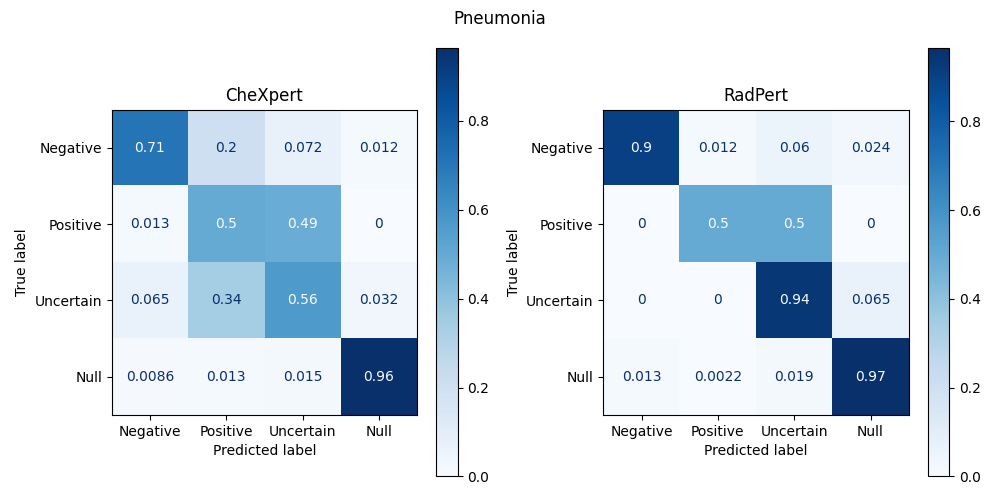}
    \end{subfigure}
    \hfill
    \begin{subfigure}{.47\textwidth}
        \centering
        \includegraphics[scale=0.26]{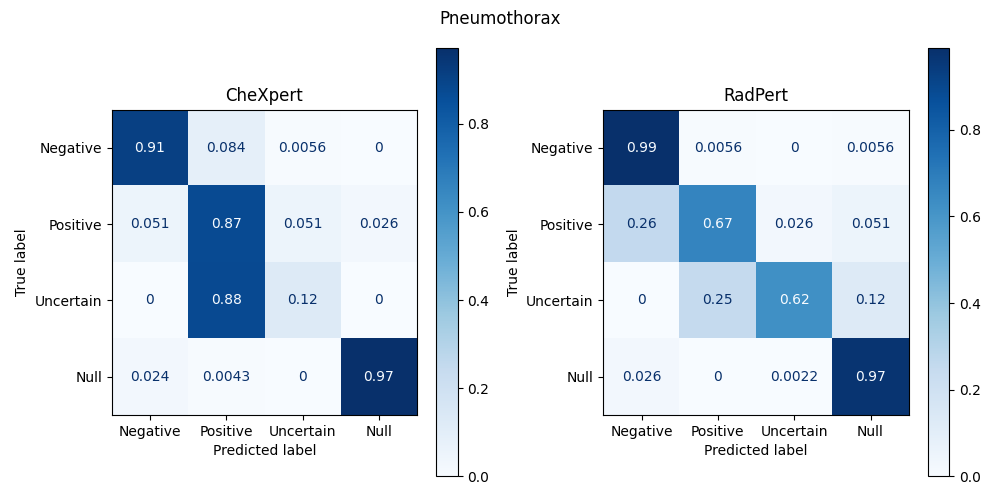}
    \end{subfigure}
    \includegraphics[scale=0.26]{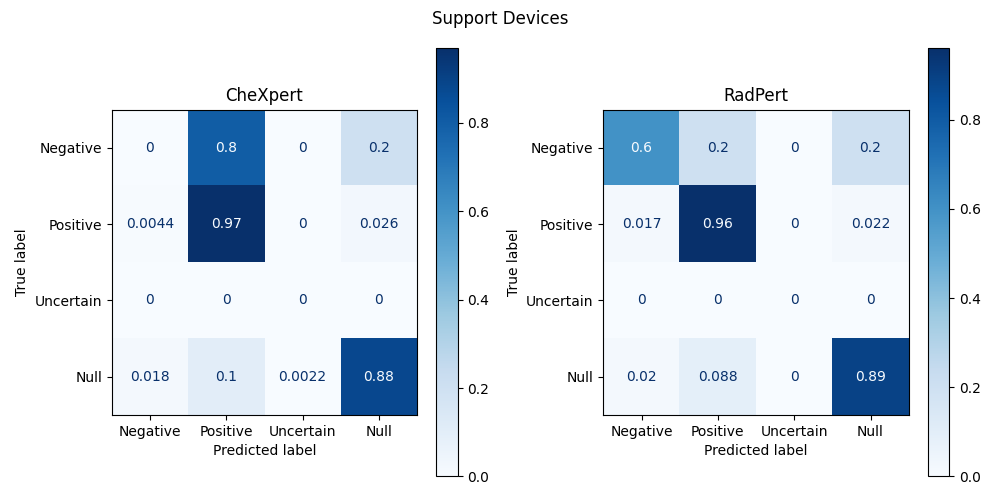}
    \caption{Normalized confusion matrices for MIMIC-CXR gold-standard test set.}
    \label{fig:conf_mimic}
\end{figure*}

\begin{figure*}[!ht]
    \centering
    \begin{subfigure}{.47\textwidth}
        \centering
        \includegraphics[scale=0.26]{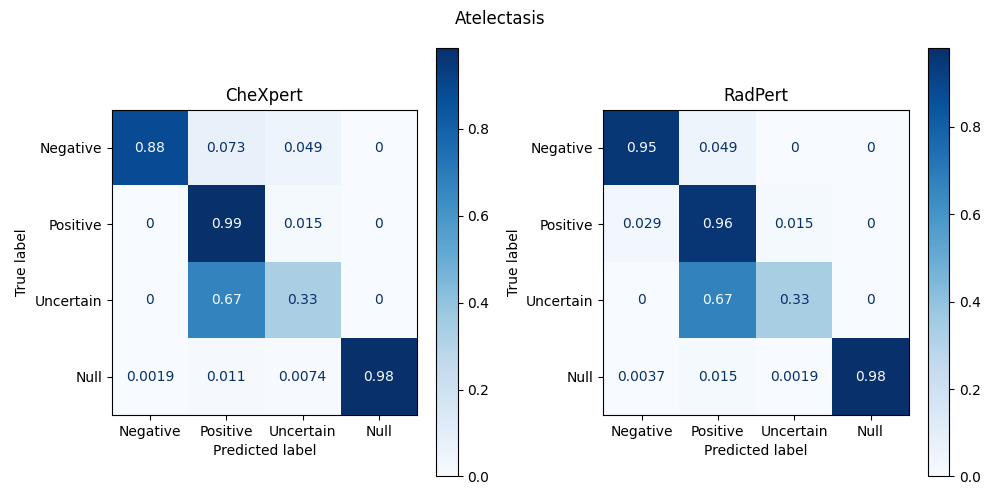}
    \end{subfigure}
    \hfill
    \begin{subfigure}{.47\textwidth}
        \centering
        \includegraphics[scale=0.26]{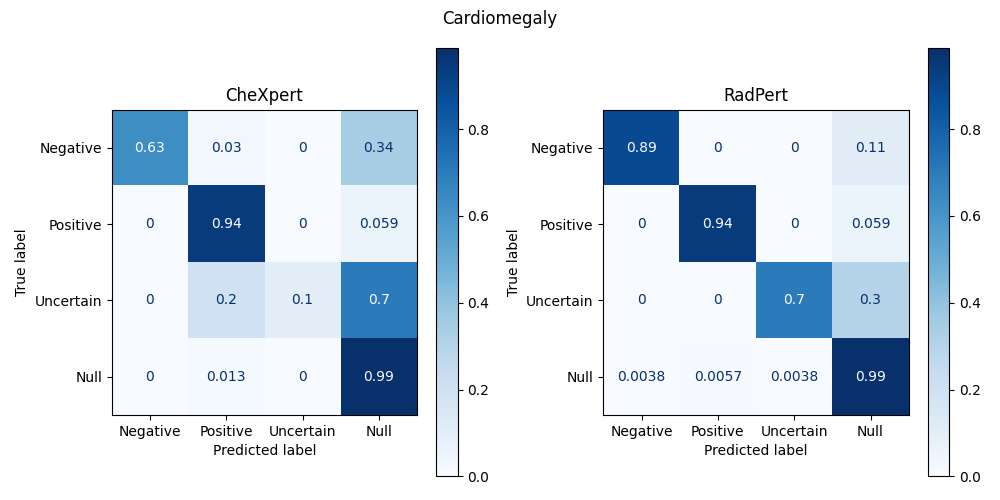}
    \end{subfigure}
    \begin{subfigure}{.47\textwidth}
        \centering
        \includegraphics[scale=0.26]{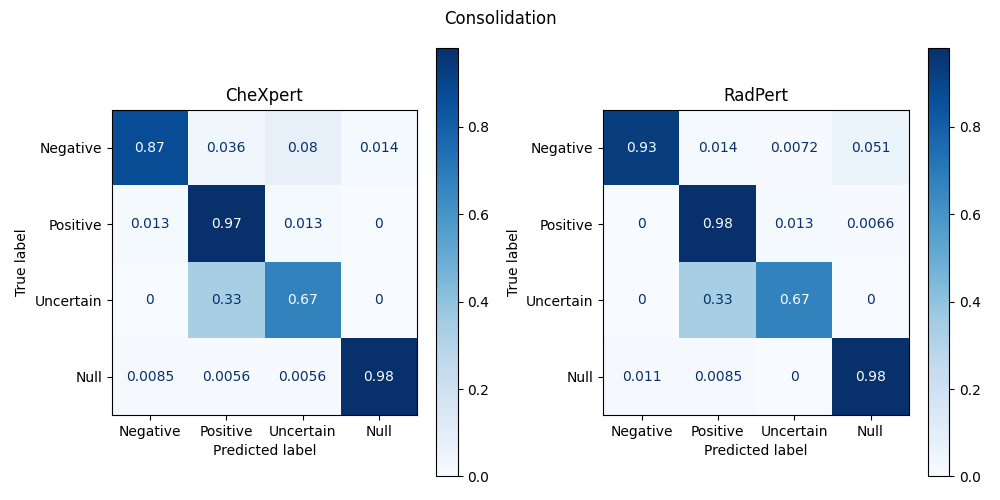}
    \end{subfigure}
    \hfill
    \begin{subfigure}{.47\textwidth}
        \centering
        \includegraphics[scale=0.26]{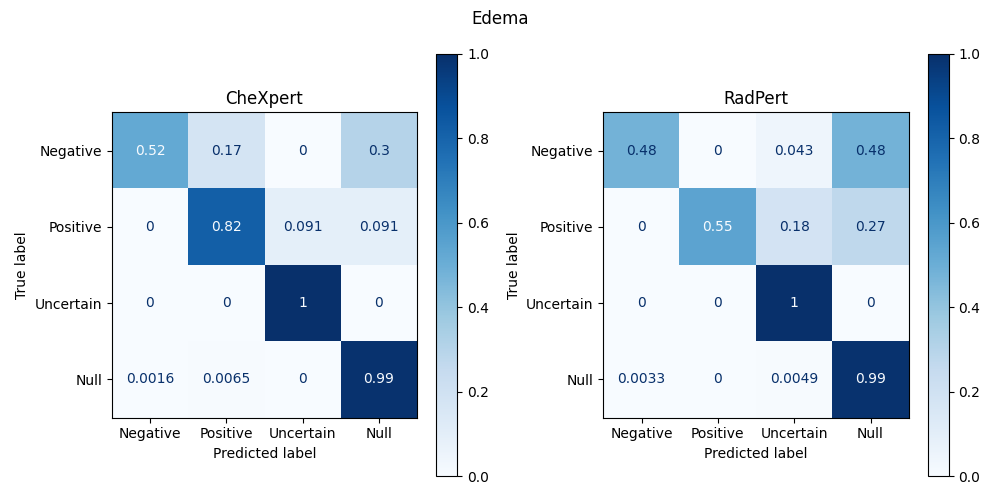}
    \end{subfigure}
    \begin{subfigure}{.47\textwidth}
        \centering
        \includegraphics[scale=0.26]{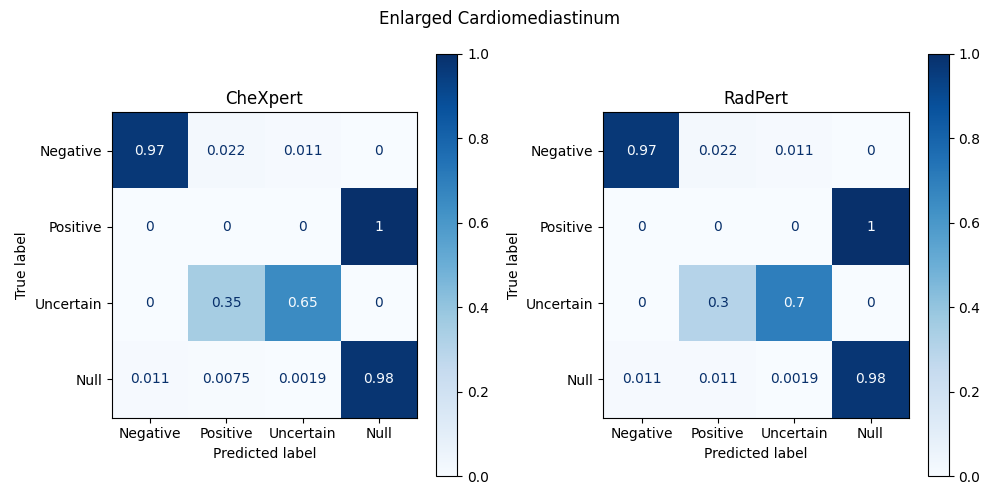}
    \end{subfigure}
    \hfill
    \begin{subfigure}{.47\textwidth}
        \centering
        \includegraphics[scale=0.26]{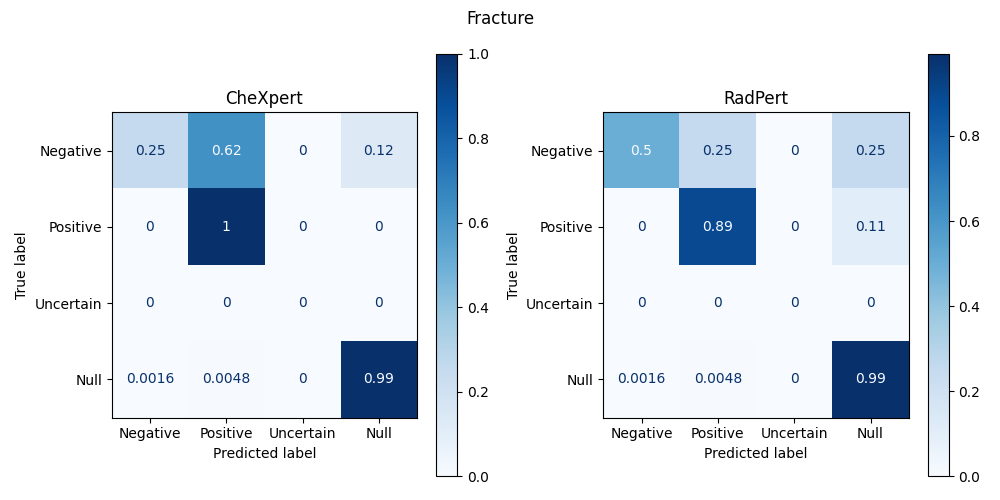}
    \end{subfigure}
    \begin{subfigure}{.47\textwidth}
        \centering
        \includegraphics[scale=0.26]{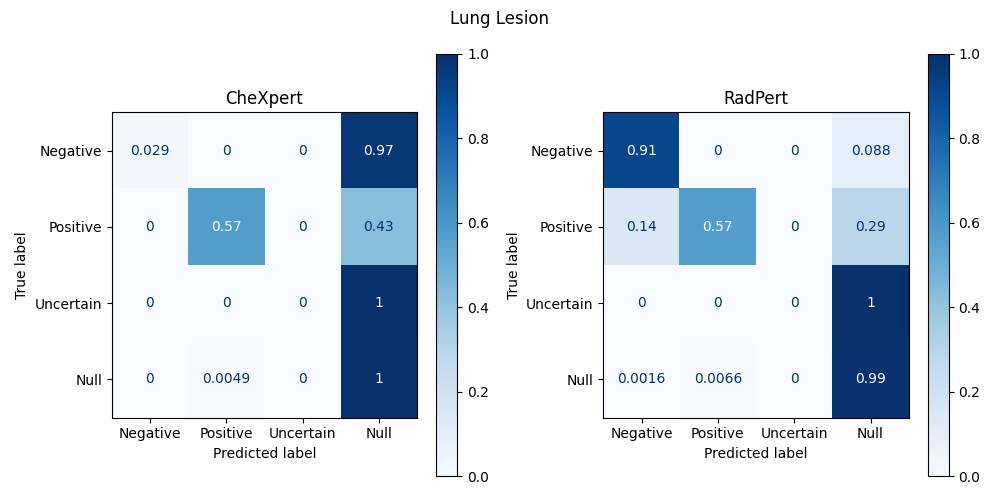}
    \end{subfigure}
    \hfill
    \begin{subfigure}{.47\textwidth}
        \centering
        \includegraphics[scale=0.26]{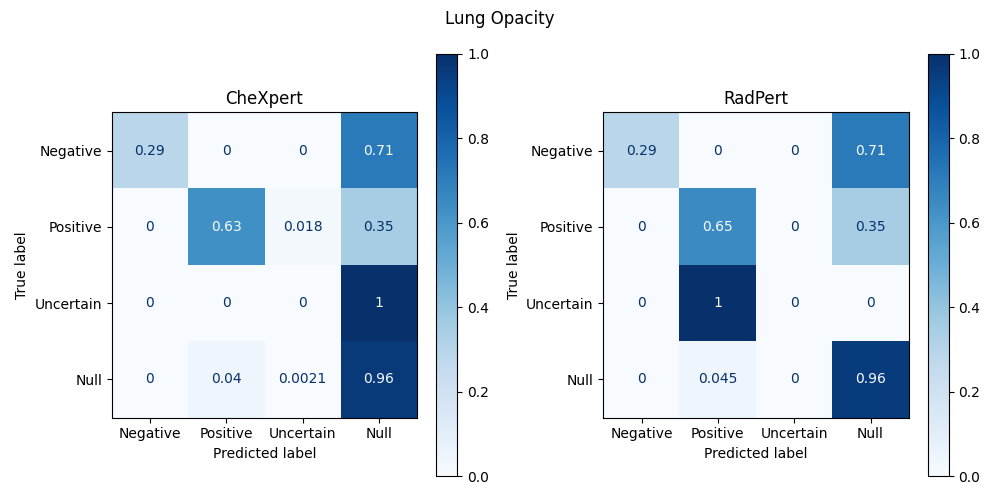}
    \end{subfigure}
    \begin{subfigure}{.47\textwidth}
        \centering
        \includegraphics[scale=0.26]{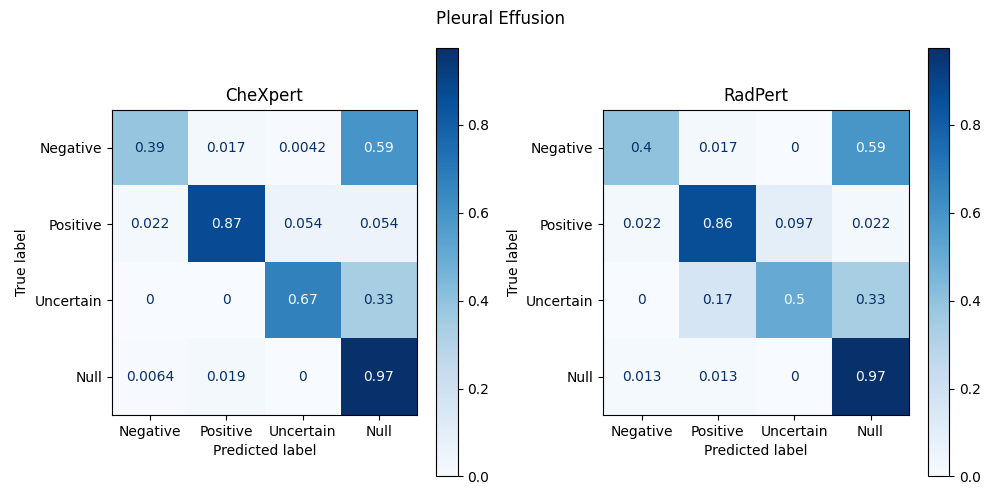}
    \end{subfigure}
    \hfill
    \begin{subfigure}{.47\textwidth}
        \centering
        \includegraphics[scale=0.26]{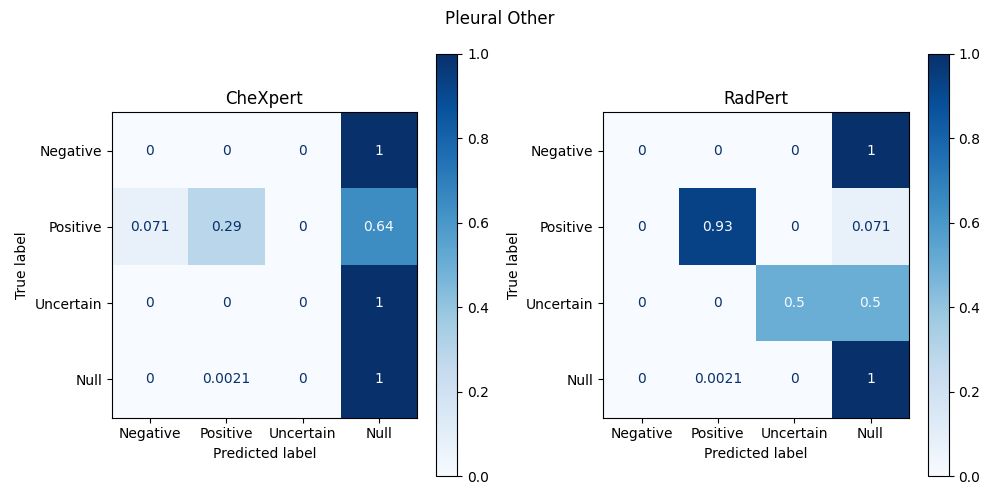}
    \end{subfigure}
    \begin{subfigure}{.47\textwidth}
        \centering
        \includegraphics[scale=0.26]{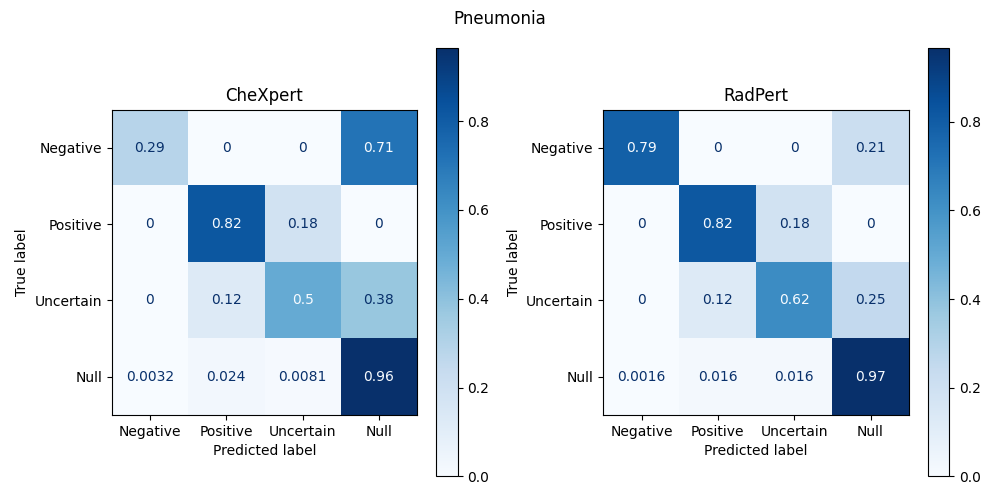}
    \end{subfigure}
    \hfill
    \begin{subfigure}{.47\textwidth}
        \centering
        \includegraphics[scale=0.26]{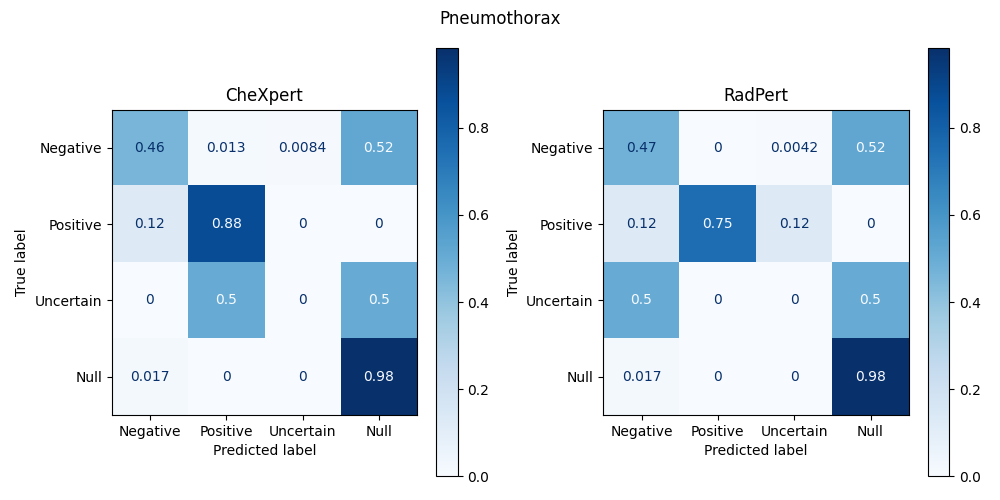}
    \end{subfigure}
    \includegraphics[scale=0.26]{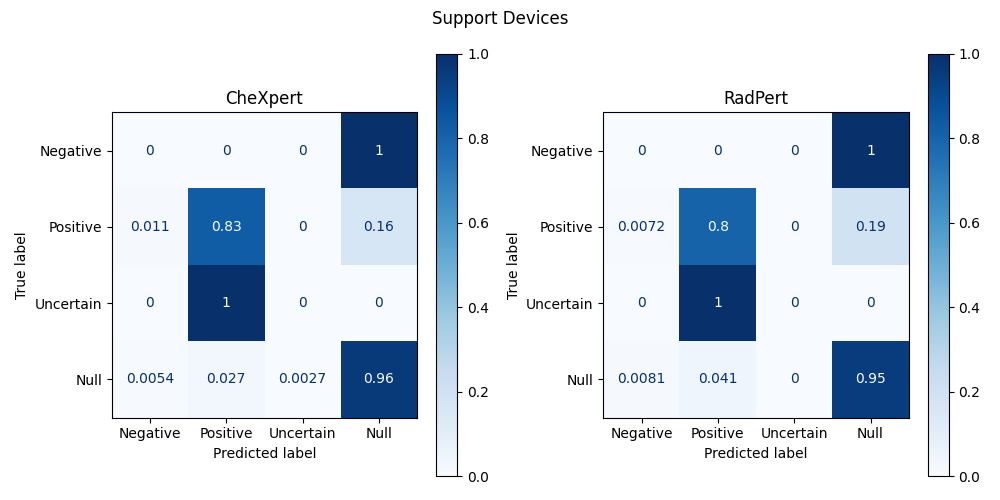}
    \caption{Normalized confusion matrices for CUH test set.}
    \label{fig:conf_cuh}
\end{figure*}

\end{document}